\documentclass[sigconf, dvipsnames, nonacm]{acmart}

\usepackage{amssymb}
\usepackage{amsmath}
\usepackage{multirow}
\usepackage[ruled, boxed]{algorithm2e}
\usepackage{comment}
\usepackage[normalem]{ulem}
\usepackage{xcolor}
\usepackage{hyperref}
\usepackage{balance}

\newcommand\XP[1]{\textcolor{black}{#1}}

\AtBeginDocument{%
  \providecommand\BibTeX{{%
    \normalfont B\kern-0.5em{\scshape i\kern-0.25em b}\kern-0.8em\TeX}}}

\setcopyright{none}
\begin{document}

\title{Semi-supervised Crowd Counting {via Density Agency}}
\titlenote{This is the accepted version of \cite{lin2022densityAgent}, the Proceedings of the 30th ACM International Conference on Multimedia (MM ’22), October 10–14, 2022, Lisboa, Portugal, DOI https://dl.acm.org/doi/10.1145/3503161.3547867, ISBN 978-1-4503-9203-7/22/10. \\ Please cite the final published version. \\ You may also refer to https://github.com/LoraLinH/Semi-supervised-Crowd-Counting-via-Density-Agency for the implementation code.}

%\title{With Help from Density Agent: Semi-supervised Crowd Counting via Contrastive Learning}


\author{Hui Lin}
\email{linhuixjtu@gmail.com}
\affiliation{%
  \institution{School of Cyber Science and Engineering\\  Xi'an Jiaotong University  PRC}
  \country{}
  %\country{P. R. China}
}

\author{Zhiheng Ma}
\email{zh.ma@siat.ac.cn}
\affiliation{%
  \institution{Shenzhen Institute of Advanced Technology\\  Chinese Academy of Science PRC}
  \country{}}

\author{Xiaopeng Hong}
\authornote{Corresponding Author}
\orcid{0000-0002-0611-0636}
\email{hongxiaopeng@ieee.org}
\affiliation{%
  \institution{School of Computer Science and Technology\\  Harbin Institute of Technology PRC} 
  \country{}
}

\author{Yaowei Wang}
\email{wangyw@pcl.ac.cn}
\affiliation{%
  \institution{Peng Cheng Laboratory PRC}
  \country{}}

\author{Zhou Su}
\email{zhousu@ieee.org}
\affiliation{%
  \institution{Xi'an Jiaotong University PRC}
  \country{}
}

\renewcommand{\shortauthors}{ }


\begin{abstract}
%As annotations for crowd images are laborious and time-consuming due to the dense crowd scenarios, semi-supervised crowd counting has attracted growing attention.
In this paper, we propose a new agency-guided semi-supervised counting approach. First, we build a learnable auxiliary structure, namely the density agency to \XP{bring the recognized foreground regional features close to} corresponding density sub-classes (agents) and push away background ones. Second, we propose a density-guided contrastive learning loss to consolidate the backbone feature extractor. Third, we build a regression head by using a transformer structure to refine the foreground features further. Finally, an efficient noise depression loss is provided to minimize the negative influence of annotation noises. Extensive experiments on four challenging crowd counting datasets demonstrate that our method achieves superior performance to the state-of-the-art semi-supervised counting methods \XP{by a large margin}. \href{https://github.com/LoraLinH/Semi-supervised-Crowd-Counting-via-Density-Agency}{Code} is available.

%As annotations for crowd images is laborious and time-consuming due to the dense crowd scenarios, semi-supervised crowd counting has attracted a growing attention. In this paper, we propose a new semi-supervised counting approach from the perspective of building a sufficient and reliable supervision among all images. First, we build a density agency based on foreground similarity to measure the correlations of features. Second, we propose a density guided contrastive learning structure to filter unreliable supervisions. Third, we build a regression head which separates the background and focuses on foreground refinement. And finally, an efficient noise depression loss is provided to minimize the negative influence by annotation noises. Extensive experiments on four challenging crowd counting datasets demonstrate that our method achieves superior performance to the state-of-the-art semi-supervised methods.

\end{abstract}

\begin{CCSXML}
<ccs2012>
   <concept>
       <concept_id>10010147.10010178.10010224</concept_id>
       <concept_desc>Computing methodologies~Computer vision</concept_desc>
       <concept_significance>500</concept_significance>
       </concept>
   <concept>
       <concept_id>10010147.10010257.10010282.10011305</concept_id>
       <concept_desc>Computing methodologies~Semi-supervised learning settings</concept_desc>
       <concept_significance>300</concept_significance>
       </concept>
   <concept>
       <concept_id>10010147.10010257.10010282.10010292</concept_id>
       <concept_desc>Computing methodologies~Learning from implicit feedback</concept_desc>
       <concept_significance>300</concept_significance>
       </concept>
 </ccs2012>
\end{CCSXML}

\ccsdesc[500]{Computing methodologies~Computer vision}
\ccsdesc[300]{Computing methodologies~Semi-supervised learning settings}
\ccsdesc[300]{Computing methodologies~Learning from implicit feedback}

\keywords{Crowd Counting, Semi-supervised, Contrastive Learning}

\maketitle

\section{Introduction}
Crowd counting~\cite{lin2022boosting, ma2021towards, lin2021object, wang2022eccnas} is to estimate the number of people in an image that could be very crowded. Since its wide applications in crowd surveillance, congestion estimation and public security, it has gained considerable attention in recent years. However, a stable and accurate crowd counter often relies on \XP{sufficient labeled images, where} annotations are marked at every head centers. The annotation process can be labor-intensive and time-consuming. For instance, UCF-QNRF~\cite{idrees2018composition} dataset involves 1.25 million annotations, which takes 2,000 human hours in total.
Therefore, \XP{semi-supervised crowd counting attempts} to reduce the annotation cost while achieving good counting accuracy. \XP{It takes advantage of and extracts extra knowledge from unlabeled data, which can be obtained cheaply.}
%Since crowd images are much easier and cheaper to access, we will extract knowledge from unlabeled data and take advantage of it.

\begin{figure}[t]
\begin{center}
    \includegraphics[width=0.43\textwidth]{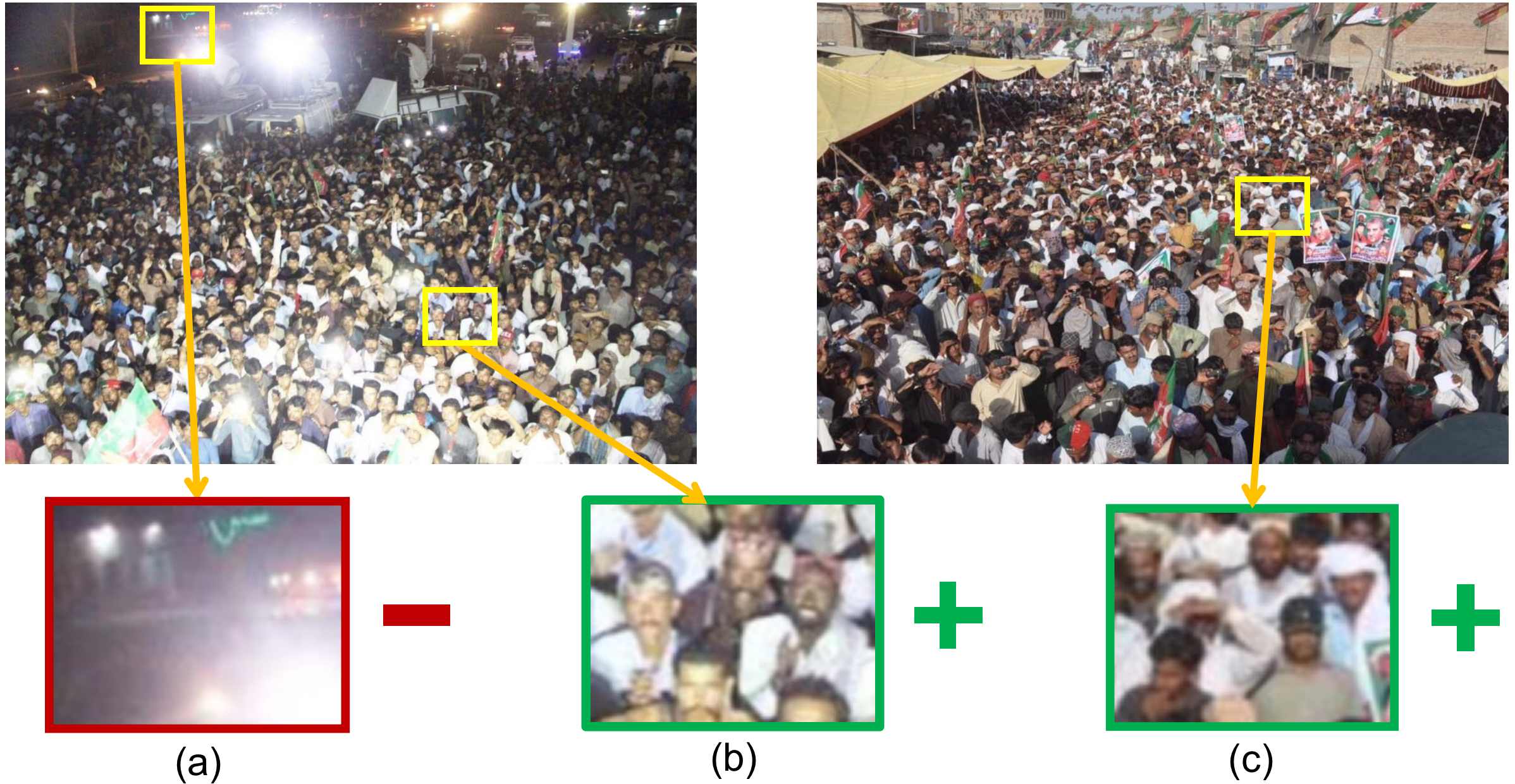}
\end{center}
\vspace{-3mm}
\caption{{The density cues can be explored by correlating image regions of different count levels}. Although the regions (a/b) are cropped from the same image, their semantics, i.e. the densities represented by them are quite different. Instead, we can find (c) similar to the foreground (b) in other images.}
\label{fig:introduction}
\vspace{-5mm}
\end{figure}

{Previous semi-supervised counting methods rely on an extra  auxiliary task like binary segmentation~\cite{meng2021spatial, liu2020semi} \XP{and} count ranking~\cite{liu2018leveraging, liu2019exploiting} to (self-) supervise the counting network in a multi-task paradigm. They usually generate pseudo-labels using methods like Gaussian process~\cite{sindagi2020learning} and teacher-student models~\cite{meng2021spatial}. These methods can effectively reduce the annotation cost. Nonetheless, the performance is far from perfect. On the one hand, self-supervised learning leveraging rules on isolated images may ignore rich information among different images. On the other hand, it is difficult to produce high-quality pseudo-labels on top of the coarse, image-level similarity, which are calculated through neural nets trained on a small amount of labeled images.}
%Previous semi-supervised counting methods can be briefly categorized into two types. The first type generates pseudo-labels based on Gaussian process~\cite{sindagi2020learning} or teacher model~\cite{meng2021spatial} to supervise the learning of unlabeled data. The second type leverages on some intrinsic rules about density to achieve self-supervision, such as containment relationship~\cite{liu2018leveraging} or ranking of density thresholds~\cite{liu2020semi}. However, these supervisions are limited within one image, while the supervision signals among different data are mutually exclusive. We believe that this isolated supervision of unlabeled data is the main factor limiting the counting potential, since the information cannot propagate between different images.

%On the one hand, the isolated supervision is unreliable, for it is difficult for pseudo-labels to achieve high accuracy with a small amount of labeled data. Under such noisy and coarse pseudo-label supervisions, the performance of the model will deteriorate by this unreliability. On the other hand, the supervision signals are also insufficient. As the self-supervised density criteria are often simple and intuitive, the model can easily satisfy these rules and thus has limited impact on improving the overall counting accuracy.

To address the limitations, {in this paper, we propose a new semi-supervised counting framework for reliable and sufficient supervision from limited labels. The novelty of this paper comes from a density agency mechanism for semi-supervised counting modeling.}

{The rationale of the density agency arises from the fact that the density cues can be mined by comparing image regions. An example is shown in Figure~\ref{fig:introduction}. When we compare a background region (a) and a foreground one (b), though the two regions are from the same image, they appear quite different. On the other hand, for a foreground region (b), we can easily find its apparent counterpart with a similar count like (c) in other crowd images. As there are huge numbers or even infinite background regions like (a), it is impossible to model them all. A more plausible alternative is to model foreground instances in line with different count levels.}

On this basis, we propose the density agency guided semi supervised crowd counting model \XP{to associate the regions of different images for obtaining reliable supervised signals}. The \emph{density agency} is a learnable auxiliary structure associated with the counting backbone network. It is made up of a group of density-level agents and an agent allocator, and
% are learnt only using the features of \emph{labeled} regions to provide reliable supervised signals for unlabeled data.
acts as an intermediary to identify whether an instance (an image region) is foreground or not. For foreground instances, it further quantifies the density levels of foreground instances, subdivides the foreground instances (regions with crowds) into sub-classes according to their density levels, and assigns corresponding density-level agents. It then measures the correlations between the instances and agents, which plays a significant role in guiding the learning process of the crowd counter. Generally speaking, during training, the density agency \XP{brings} the recognized foreground instances to the corresponding sub-classes (density-level agents) and pushes away negative instances in feature space. {More specifically, we design an efficient learning method to optimize the agents. We then propose a density-level guided contrastive loss to provide fine-grained signals for consolidating the learning of feature extractors. We further introduce a transformer structure to bridge the feature extractor and the density estimation module and refine the recognized foreground features. Finally, we devise a noise depression Bayesian loss to eliminate the annotation noises.}

%\XP{We realize these functions by designing a semi-supervised learning paradigm adjoint to the dense agency. More specifically, we propose a density-level guided contrastive loss for fine-grained self-supervision. Moreover, we also devise a noise depression Bayesian loss to further eliminate the annotation noises.} 
%The proposed model is named Density Agency based Crowd Counting (DACount).

%\XP{We evaluate the proposed xxxx on xxxx. ...}

We evaluate the proposed model on four challenging crowd counting datasets, i.e. ShanghaiTech A and B~\cite{zhang2016single}, UCF-QNRF~\cite{idrees2018composition} and JHU-Crowd++~\cite{sindagi2020jhu}. The experimental results show that no matter what percentage of data is labeled, our model outperforms state-of-the-art semi-supervised counting methods by a large margin. {For example, under the most challenging setting of only 5\% labeled data on the QNRF dataset, our method sets up new state-of-art accuracy and reduces the MAE from 160.0 to 120.2, achieving a \XP{notable about 25\% reduction in mean absolute error.}}

In summary, {we propose a novel semi-supervised crowd counting model, named Density Agency based Crowd Counting (DACount), with the following main contributions}:

\begin{itemize}
    \item We introduce a density agency guided semi-supervised learning scheme, which is able to construct sufficient supervisions for unlabeled data, breaking the supervision border between images.
    
    \item We design a density-guided uncertainty-aware contrastive loss {based on} the semantic differences of foreground features.
    %and a noise aware filter mechanism to improve reliability.
    
    \item {We design a noise depression Bayesian loss to alleviate the negative influence of annotation noise.}
    %We design a foreground-refined transformer to further \XP{generate enhanced features for density estimation by considering the correlations and differences among foreground background features.
    
	%\item A noise depression Bayesian loss is proposed to minimize the negative influence of annotation noises.
	
	\item {We set up new state-of-the-art performance for semi-supervised crowd counting on popular benchmarks.}
	%We conduct extensive experiments on popular benchmarks and validate that DACount outperforms state-of-the-art semi-supervised algorithms by large margins.
	
\end{itemize}

\begin{figure*}[t]
\begin{center}
    \includegraphics[width=0.95\textwidth]{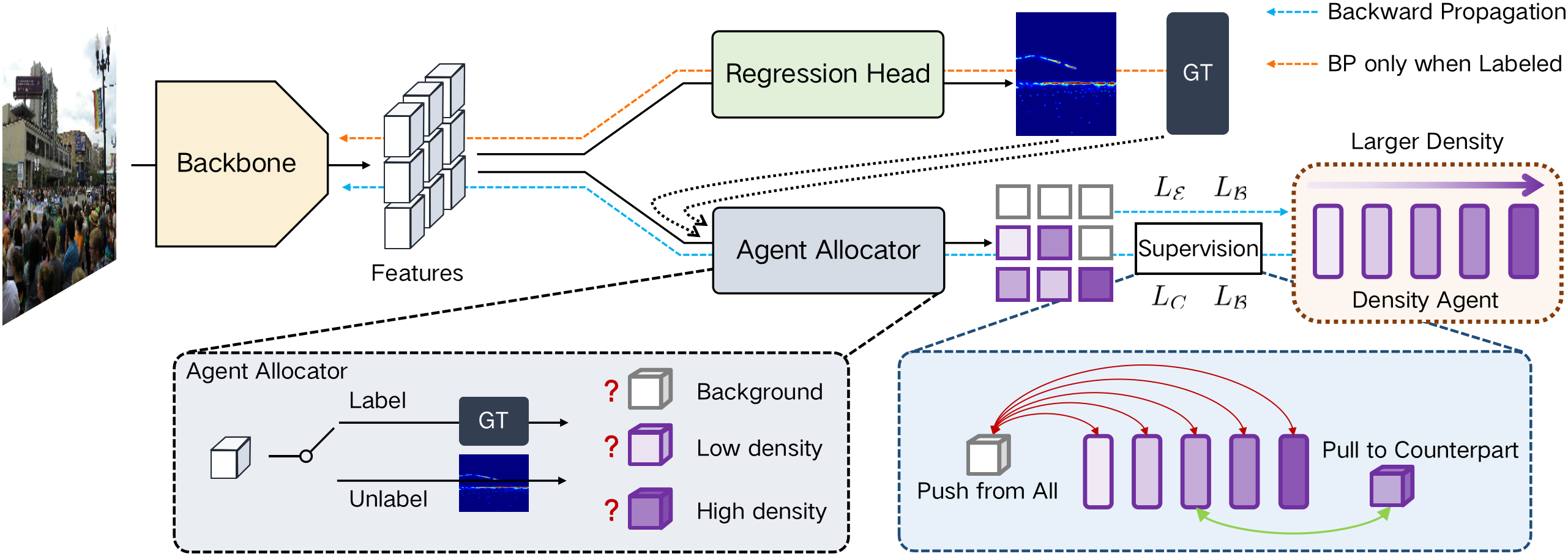}
\end{center}
\vspace{-3mm}
% \caption{Details of our counting \XP{framework}. It consists of a CNN backbone to extract features, a foreground predictor to segment foreground and background, a transformer to refine foreground features and a density estimator to generate the density map. Except for the backbone, the training of the remaining three modules only relies on accurate labeled data. Throughout the semi-supervised training, the learnable density agent is used to separate foreground and background features, and endow the density information into different foreground features.}
\caption{Illustration of our counting framework and the training pipeline of the density {agency}. Features extracted by the backbone will be divided into background and different subclasses of foreground by the agent allocator. Then the agents directly supervise the backbone at intermediate feature layers instead of predicted maps by pushing all background farther and {pulling} the right foreground pairs closer. We \XP{design specific} loss functions to update the density agents (Eq.~\ref{eq:foreground} and Eq.~\ref{eq:background}) and the counting model (Eq.~\ref{eq:contrastive} and Eq.~\ref{eq:background}) respectively. The direction of the blue arrow represents the back-propagation update process of the loss gradients.}
\label{fig:structure}
\vspace{-3mm}
\end{figure*}

\section{Related Works}

\subsection{Fully-supervised Crowd Counting}
In recent years, fully-supervised crowd counting has experienced a rapid development. With the help of deep CNN, the multi-column model~\cite{zhang2016single} is proposed to regress the prediction to the pseudo density map generated with adaptive Gaussian kernels. \XP{CSRnet uses dilated kernels} to perform accurate count estimation of highly congested scenes~\cite{li2018csrnet}.
%\cite{li2018csrnet} introduces CSRNet, which uses dilated kernels to perform accurate count estimation of highly congested scenes.
In \cite{cao2018scale}, a scale aggregation network is proposed to exploit the local correlation and structural similarity. Ma \emph{el al.} adopt Bayesian assumption and introduces Bayesian Loss (BL) to calculate the expected count of pixels~\cite{ma2019bayesian}. \XP{Wang \emph{el al.} propose to use optimal transport to match the distributions of the point annotations and the density maps~\cite{wang2020distribution}.} In \cite{ma2021learning}, the unbalanced optimal transport is introduced to quantify the discrepancy between predicted density maps and point annotations. \XP{S3} further proposes semi-balanced Sinkhorn divergence to solve the limitations of amount constraint and entropic bias~\cite{lin2021direct}. In addition, methods based on multi-scale mechanisms~\cite{zeng2017multi,sindagi2019multi,ma2020learning}, segmentation~\cite{sajid2016crowd} and perspective estimation~\cite{shi2019revisiting,yan2019perspective} are proposed to overcome the limitations of perspective and scale variations in crowd images. Moreover, BinLoss~\cite{shivapuja2021wisdom}, P2PNet~\cite{song2021rethinking}, UEPNet~\cite{wang2021uniformity} and MFDC~\cite{liu2021exploiting} further propose new solutions for crowd counting. Lately, the development of Transformer~\cite{vaswani2017attention} also boosts counting performance. \XP{The study}~\cite{wei2021scene} learns adaptive feature representations with a deformable attention in Transformer network. \XP{In \cite{lin2022boosting},} a learnable local region attention with a corresponding regularization is proposed to address large-scale variations in crowd images.

\subsection{Semi-/Weakly supervised Crowd Counting}
As \XP{labeling crowd images} is time-consuming and labor-intensive, researchers gradually focus on semi-/weakly supervised learning to find low-cost solutions for crowd counting. \XP{L2R} introduces a crop ranking loss by learning containment relationships to exploit unlabeled images~\cite{liu2018leveraging}. Meanwhile, \XP{its extension} improves the ranking strategy and regards it as a self-supervised proxy task~\cite{liu2019exploiting}. Yang \emph{et al.} \cite{yang2020weakly} propose a soft-label sorting network and only uses the crowd number as supervision. \XP{The study} \cite{sindagi2020learning} proposes a Gaussian Process-based iterative learning mechanism to generate pseudo-ground truth for unlabeled data. Furthermore, \XP{IRAST} leverages on a set of inter-related binary segmentation tasks for self-training to exploit the underlying constraints of unlabeled data~\cite{liu2020semi}. The spatial uncertainty aware teacher-student framework is proposed to alleviate the noisy supervision from unlabeled data~\cite{meng2021spatial}.

Our method is distinct to previous semi-supervised counting approaches. \XP{The methods ~\cite{sindagi2020learning, meng2021spatial} generate pseudo-labels based on image-level similarity, while our method takes direct advantage of the region-level features. Moreover, compared to self-supervision methods which usually work with a single image~\cite{liu2018leveraging, liu2019exploiting, liu2020semi}, we make use of much richer supervision signals among different images.}
%For methods generated pseudo-labels~\cite{sindagi2020learning, meng2021spatial}, our method is based on feature level and avoids coarse, image-level similarity matching. And compared to methods leveraged on self-supervision~\cite{liu2018leveraging, liu2019exploiting, liu2020semi}, we make full use of the information among different crowd images and propagate the supervision signals across images.

\subsection{Contrastive Learning}
Contrastive learning can be formally defined by creating positive and negative data pairs, which are semantically similar and dissimilar respectively. The goal is to pull positive pairs together and push negative pairs apart in the feature space~\cite{hadsell2006dimensionality}. It has been widely used in self-supervised representation learning~\cite{chen2020improved, chen2020simple, he2020momentum, sermanet2018time}, where the contrastive loss serves as an unsupervised objective function to measure the similarities of sample pairs. Previous works also apply contrastive learning into various vision tasks, such as object detection~\cite{xie2021detco, sun2021fsce, wei2021aligning, wang2021dense} and semantic segmentation~\cite{wang2021exploring, alonso2021semi, zhao2021contrastive}. These tasks inherently have object/class level positive and negative samples/features, which suit for modeling the contrast. Recently, some works also adopt contrastive idea in semi-supervised learning~\cite{li2021comatch,singh2021semi,alonso2021semi,jiang2021guided}.

\XP{The objective of contrastive learning can be formulated in various functions, including max-margin contrastive loss~\cite{hadsell2006dimensionality}, triplet loss~\cite{weinberger2009distance}, and angular loss~\cite{wang2017deep}. In this work, we mainly consider the InfoNCE loss~\cite{oord2018representation} and revisit contrastive learning in the context of the semi-supervised dense regression problem.} We design a density guided agent and mine the relationships between positive and negative features, \XP{which yields improved performance.}
%We adopt the idea of contrastive learning into semi-supervised dense regression problem (Crowd Counting). We design a density guided agent and mine the relationships between positive and negative features.

\section{The Proposed Method}

DACount is a combination of four novel modules to semi-supervised crowd counting:  a learnable density agency, a self-supervised contrastive loss, a foreground transformer and a noise depression loss. In this section, we first \XP{introduce the notations for} the semi-supervised counting problem and then introduce a density agency to establish a shared supervision \XP{scheme} among all \XP{training} images. It consists of an agent allocator and a set of density agents. On that basis, we propose the \XP{density agent guided} contrastive \XP{learning loss}. Then we \XP{elaborate} the proposed network structure with a transformer to \XP{rectify} foreground features. Finally, a noise depression Bayesian loss \XP{is} introduced to enhance the update of regression head under few labeled data.

% \subsection{Notations and Preliminaries}
\subsection{Density Agent}\label{sec:density agent}

Assume we have a labeled dataset $\mathcal{L} = \{(x^i,y^i)\}^{N_l}_{i=1}$ of $N_l$ labeled samples, where $x^i$ is a crowd image and $y^i$ is the corresponding point ground truth. An unlabeled dataset can also be denoted as $\mathcal{U} = \{(x^i)\}^{N_u}_{i=1}$ of $N_u$ unlabeled samples, where the size of it is usually much larger than that of the $\mathcal{L}$, i.e., $N_l << N_u$.

In crowd counting, as shown by the example in Figure~\ref{fig:introduction}, we discover that the \XP{region-level} variations among different images are mainly from background,
\XP{while regions with crowds may have similar appearance.}
%instead of foreground (regions with crowd).
%As shown in Figure~\ref{fig:introduction}, although the background (a) and foreground (b) are cropped from the same image, both their semantics and appearances are quite different. Instead, a similar foreground counterpart (c) can be easily found in other images.
Therefore, based on this \XP{observation}, we propose to build a density agency targeted on foreground across all images. The agency is a learnable auxiliary structure associated with the counting model, acting as an intermediary to explore the correlations of foreground from different images and background in feature space. Specifically, the agent consists of a group of density-level agents and an agent allocator. 
%It consists of an agent locator to divide features and density agents to guide supervisions.

{The \textbf{agent allocator} is designed to divide features extracted by the backbone into {background and} different groups {of foreground} \XP{features} in line with the quantitative levels of their density and assign \textbf{density agents} accordingly. It is formulated as follows. Suppose the density range is partitioned into $N_a$ intervals, the borders of which are denoted by a series, i.e.,  $ \displaystyle (0, v_{1}, v_{2},\ldots, v_{N_a-1})$. We have $N_a$ density-level agents $\mathcal{A} = \{f_i\}^{N_a}_{i=1}$, $f_i \in \mathbb{R}^{c}$, {where $c$ is the channel dimension}. The agents can be assigned by the following semantic-driven rule.}
%Given the size $N_a$, we denote the density-level agents as $\mathcal{A} = \{(f_i)\}^{N_a}_{i=1}$, $f_i \in \mathbb{R}^{d}$. Each $f$ represents a unique density value partition. Assuming we have $N_a-1$ density interval borders, then the represented pairs can be written by
\begin{equation}
    f_1 \leftarrow \XP{(}0, v_1),\ f_2 \leftarrow [v_1, v_2),\ ...\ ,\ f_{N_a} \leftarrow [v_{N_a-1}, +\infty).\nonumber
\end{equation}
%The agent locator is used to divide features extracted by the backbone into different subclasses of foreground or background. 
\XP{The left-bottom rectangle in Figure~\ref{fig:structure} illustrates how the agent allocator maps the regions from an input image to the corresponding agents. For} labeled data, the agent allocator will leverage on the ground-truth to quantify the density levels and classify different regions. \XP{For unlabeled data, there is no ground truth available. To obtain the region-wise density levels, the unlabeled image first go through the backbone and the regression head, and then the agent allocator} use the predicted density map as the reference.

{More specifically, let  $\mathcal{E}$ and $\mathcal{B}$ denote the {foreground} regional feature set and the background regional feature set respectively. $\mathcal{E} \cap \mathcal{B} = \varnothing,\ \mathcal{E} \cup \mathcal{B} = \mathcal{F}$, where $\mathcal{F}$ denotes all region features extracted by the backbone. For a {foreground} feature $e \in \mathcal{E}$, if the density value of $e$ is in the range $[v_i, v_{i+1})$, we find its agent $f_{i} = Z(e)$ for that interval through the agent allocator.}

\XP{With the agency mechanism, we expect to pull the foreground regional features close to their density agents for the corresponding density intervals and push the background features away from all the agents. As a result, an agent is desired to uniquely represent a foreground density interval and be significantly different to the regional background features. To achieve this goal, we design a simple yet efficient agent learning algorithm, which is shown in the right-bottom part of Figure~\ref{fig:structure}.}
%The training \XP{process of the density agents} is shown in the right-bottom part of Figure~\ref{fig:structure}. 

%The goal is to supervise the foreground feature by pulling it be close to its counterpart in the density agent and the background feature by pushing it far away from all.
%Let  $\mathcal{E}$ and $\mathcal{B}$ denote the  regional feature set and the background regional feature set respectively. $\mathcal{E} \cap \mathcal{B} = \varnothing,\ \mathcal{E} \cup \mathcal{B} = \mathcal{F}$. For a feature $e \in \mathcal{E}$, suppose that the density value of $e$ is in the range $[v_i, v_{i+1})$. We find its agent $f_{i} = Z(e)$ for that interval through the agent allocator, i.e.,  Then the goal is to maximize the similarity between each pairs.

\XP{Given  $\mathcal{E}$ and $\mathcal{B}$, we optimize the agents to maximize the similarity between positive pairs, each of which is formed of a foreground feature vector and its density agent and minimize the one between negative pairs, each of which contains a background feature vector and any agents. In details, the process is formulated as follows.}

\begin{equation}\label{eq:learn_agents}
    {\min \left (L_{\mathcal{E}}+ L_{\mathcal{B}}  \right )},
\end{equation}
% \begin{equation}\label{eq:learn_agents}
%     \XP{\mathcal{A}^* = \arg \min_{\mathcal{A}}  L_{\mathcal{\mathcal{A}}} = \arg \min_{\mathcal{A}}  \left (L_{\mathcal{E}}+ L_{\mathcal{B}}  \right )},
%     %\XP{\mathcal{A}^* = \arg \min_{\mathcal{A}}  L_{\mathcal{E}}+ L_{\mathcal{B}},}
% \end{equation}
where
\begin{equation}\label{eq:foreground}
    L_{\mathcal{E}} = -\sum_{e \in \mathcal{E}} s(Z(e), e)
\end{equation}
and
\begin{equation}\label{eq:background}
    L_{\mathcal{B}} = \frac{1}{N_a} \sum_{f \in \mathcal{A}} \sum_{b \in \mathcal{B}} s(f, b).
\end{equation}
\XP{$L_{\mathcal{E}}$ is responsible for pulling close  positive pairs, i.e., $(Z(e), e)$ and $L_{\mathcal{B}}$ for pushing away any agent $f$ and a background feature $b$.} 
%Meanwhile, each background feature $b \in \mathcal{B}$ should be far away from all agents $\mathcal{A}$. That is, 

% \begin{figure}[t]
% \begin{center}
%     \includegraphics[width=0.45\textwidth]{laplace.png}
% \end{center}
% \caption{A depiction of uncertainty-aware contrastive learning. Counterparts are divided into positive and negative according to the density differences. Then contrastive pairs are reweighted with uncertainty-awareness.}
% \label{fig:contrastive}
% % \vspace{-4mm}
% \end{figure}

By taking into account the range of values, we adopt the cosine similarity between two {vectors} as the measurement.
\begin{equation}
    s(f_1, f_2) = \frac{f_1 \cdot f_2}{\Vert f_1 \Vert \Vert f_2 \Vert}.
\end{equation}

%Meanwhile, each background feature $b \in \mathcal{B}$ should be far away from all agents $\mathcal{A}$. That is, 
%\begin{equation}\label{eq:background}
%    L_{\mathcal{B}} = \frac{1}{N_a} \sum_{f \in \mathcal{A}} \sum_{b \in \mathcal{B}} s(f, b).
%\end{equation}

At every stage, we use gradient descent to update the agents in order to reach the optimal solution for the current state. The optimization process of foreground and background loss can be written by,
\begin{equation}\label{eq:gradients}
\begin{aligned}
    &f_{t+1} = f_t - \gamma (\nabla_f L_{\mathcal{E}}+\nabla_f L_{\mathcal{B}}), \textup{where}\\
    &\nabla_f L_{\mathcal{E}} = \sum_{Z(e) = f} s(f, e)\frac{f}{\Vert f \Vert^2} - \frac{e}{\Vert e \Vert \Vert f \Vert},\\
    &\nabla_f L_{\mathcal{B}} = \sum_{b \in \mathcal{B}} \frac{b}{\Vert b \Vert \Vert f \Vert} - s(f, b)\frac{f}{\Vert f \Vert^2}.\\
\end{aligned}
\end{equation}
$\gamma$ is the learning rate. For computational simplicity and efficiency, the Adam optimizer~\cite{kingma2014adam} is used to update the density agents. 

\subsection{Density \XP{Agent} Guided Contrastive Learning}\label{sec:contrastive learning}

\begin{figure}[t]
\begin{center}
    \includegraphics[width=0.45\textwidth]{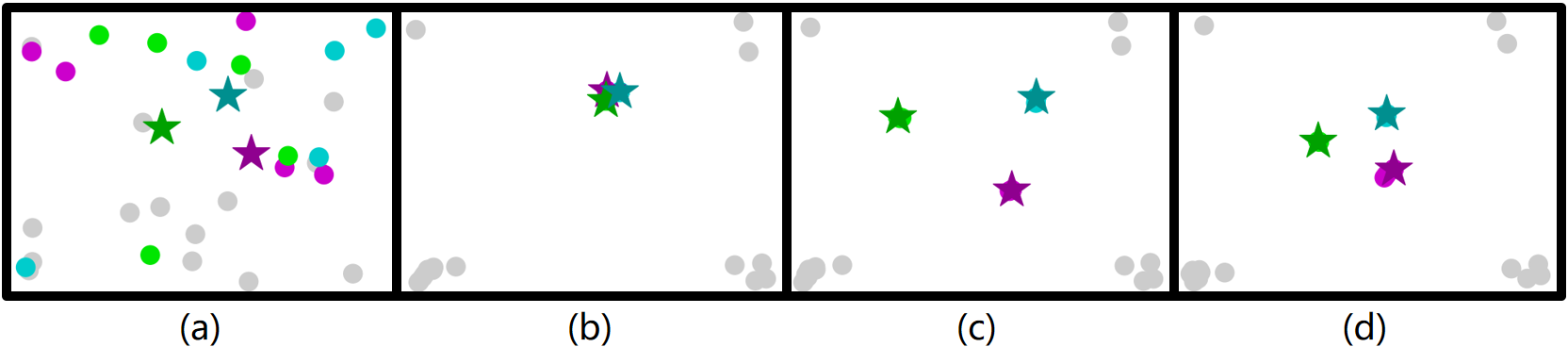}
\end{center}
\vspace{-3mm}
\caption{A toy experiment to show the influence of different supervisions. Colored and grey dots represent different foreground features and background respectively while stars represent the agents. (a) Initialization. (b) When using $L_{\mathcal{E}}$ in Eq.~\ref{eq:foreground} instead of contrastive loss $L_c$ in Eq.~\ref{eq:contrastive} {to update the counting model, the foreground features collapse to be similar.} (c) {When using $L_c$ instead of $L_{\mathcal{E}}$} to update the agency, the distance between the foreground features may be greater than that between foreground and background. (d) When using $L_{\mathcal{B}}$ in Eq.~\ref{eq:background} and $L_c$ {to update the counting model and using $L_{\mathcal{B}}$ and $L_{\mathcal{E}}$ to update the agency}, foreground features are similar enough and also clearly differentiated.}
\label{fig:toy}
\vspace{-3mm}
\end{figure}

\XP{Given a foreground feature, either labeled or unlabeled, the agent allocator assigns an agent to it. }
%In the above section, each foreground can find its counterpart based on the density matching. 
However, there are two key problems. First, \XP{as there are errors in the generated density maps (for unlabeled data), the allocator inevitably encounters uncertainties in matching agents and features. As a result, a wrong pulling will further lead to poorer density prediction}. Second, Eq.~\ref{eq:foreground} \XP{and~\ref{eq:background}} only supervise the differences between foreground and background, and does not make a finer distinction among foreground densities. As it shown in Figure~\ref{fig:toy} (b), this\XP{may} make the agent collapse to a common position in the feature space. In this section, our goal is to build a reliable scheme to first select signals from foreground supervision and then leverage on the contrastive learning to separate different foreground features.

\XP{In practice, especially in noisy cases when only partial data are labeled, the predicted density value $d$ of a foreground feature may perturb within a certain range. Put another way, this uncertainty stands for the matching probability of a foreground feature over different agents as well. The higher the {matching} probability, the more likely the foreground features can be regarded as a positive proposal, and vice versa. Based on this understanding, we assume that this uncertainty conforms to a Laplace distribution and assign higher weights to proposals that are clearly divided, and down weigh those which are on the boundary between positive and negative. As a result, suppose $d$ falls into the $i$-th interval,} we define the uncertainty-aware weight $\omega_i$ as, 

%for a foreground feature, when its density $v$ perturbs within a certain range, we assume that this uncertainty conforms to a Laplace distribution and also represents the matching probability for different agents. \XPDel{We believe that} When the matching probability is high enough, the pair can be regarded as a positive proposal, while some pairs with very low matching probabilities are regarded as negative ones. Meanwhile, we assign higher weights to proposals that are certainly divided, and reduce the weights of those which are on the boundary between positive and negative. Therefore, we define the uncertainty-aware weights as, 

%For a foreground feature, when its density $v$ perturbs within a certain range, we assume that this uncertainty conforms to a Laplace distribution and also represents the matching probability for different agents. \XPDel{We believe that} When the matching probability is high enough, the pair can be regarded as a positive proposal, while some pairs with very low matching probabilities are regarded as negative ones. Meanwhile, we assign higher weights to proposals that are certainly divided, and reduce the weights of those which are on the boundary between positive and negative. Therefore, we define the uncertainty-aware weights as, 
\begin{equation}
    \label{eq:hatomega}
        \omega_i =  8|\hat{\omega}_i - 0.25|,
    %\omega_i =  8|\hat{\omega} - 0.25|,\quad \hat{\omega_i} \sim \frac{1}{4}e^{-\frac{|a_i-v|}{2}} ,\quad a_i = \frac{1}{2} (v_{i-1} + v_i).
\end{equation}
\XP{where $\hat{\omega_i}$ is the matching probability, subject to a Laplace distribution, i.e., $\quad \hat{\omega_i} \sim \mathbf{Laplace}\left (a_i, 2 \right ) = \frac{1}{4}e^{-\frac{|d-a_i|}{2}}$. The location parameter $a_i$ is the central value of the density interval} for $f_i$. In extreme cases, $a_1=\frac{1}{2}v_1$ and $a_{N_a}=v_{N_a-1}$. \XP{The parameters in Eq.~\ref{eq:hatomega} are set to ensure the value of the weight $\omega_i$ within the range of $[0, 1]$.}
%$\hat{\omega}$ is the matching probability and $\omega$ is the weight.

\XP{In light of Eq.~\ref{eq:hatomega},} we select reliable pairs for supervision. We consider features which the matching probability above the certain threshold $0.25$ to be positive proposals and others to be negative. Matching positive pairs should be pulled in further, while selected negatives should be pushed farther, aiming to increase the dissimilarity of features with different density values. Then for feature $e$, we define the uncertainty-aware contrastive loss as
\begin{equation}\label{eq:contrastive}
    L_c(e) = -log\ \frac{\sum_{\hat{\omega}_i\geq 0.25}\omega_i u_i}{\sum_{j=1}^{N_a} \omega_j u_j}, \quad u_i = e^{s(f_i,e) / \tau}
\end{equation}
where $\tau$ is a temperature parameter. Then the final loss is the sum of contrastive loss of all foreground features and background loss with weight $\lambda_b$:
\begin{equation}\label{eq:contrastive2}
    L_C = \sum_{e \in \mathcal{E}} L_c(e) + \lambda_b L_\mathcal{B}.
\end{equation}
In experiments, we will set $\lambda_b = 1$ for simplification. 

Specifically, {we use different loss functions to update the density agency and the counting model respectively. \XP{To update the} density agents, we adopt Eq.~\ref{eq:foreground} as foreground loss and Eq.~\ref{eq:background} as background loss. The detailed learning process is described in the last few paragraphs of Section~\ref{sec:density agent}. Meanwhile, \XP{to update the} counting model, we adopt the contrastive loss in Eq.~\ref{eq:contrastive} as foreground loss and Eq.~\ref{eq:background} as background loss. This update solution is in consideration of the pace of gradient updates.} For a negative counterpart $f_i$, $\hat{\omega}_i<0.25$, its gradient of Eq.~\ref{eq:contrastive} is 
\begin{equation}
    \nabla_f L_c = \frac{1}{\tau} \cdot \frac{\omega_i u_i}{\sum_{j=1}^{N_a} \omega_j u_j} \cdot (\frac{e}{\Vert e \Vert \Vert f_i \Vert} - s(f_i, e)\frac{f_i}{\Vert f_i \Vert^2}).
\end{equation}
Comparing to the gradient of background loss in Eq.~\ref{eq:gradients}, an inappropriate negative pair with a larger similarity may cause the agent to be pushed away by a pace greater than that of the background. As shown in Figure~\ref{fig:toy} (c), the distance between the foreground may be greater than that between \XP{the} foreground and background {features} when using the contrastive loss for agent update.

\subsection{Network Structure}\label{sec:structure}

\XP{To support the above functions, a powerful and reliable regression head is essential. In this regard,}
%In the above supervisions, we need to separate features based on density. For labeled data, we will directly divide features based on ground-truth, while for unlabeled data, we will rely on the density map predicted by the counting model. Therefore, ensuring a correct regression head from features is quite essential. To ensure its reliability,
we build the semi-supervised training scheme and further introduce a transformer \XP{for refining the foreground features}.

For the training scheme, we decide to only leverage on labeled data to train the regression head in order to avoid contamination by inaccurate supervision from unlabeled data, keeping a \XP{`}pure' head. To achieve this, the unlabeled data will only act directly on the feature extractor, instead of the whole model. Therefore, when using the predicted density map to divide features, we will stop gradients of the regression head.

\begin{figure}[t]
\begin{center}
    \includegraphics[width=0.48\textwidth]{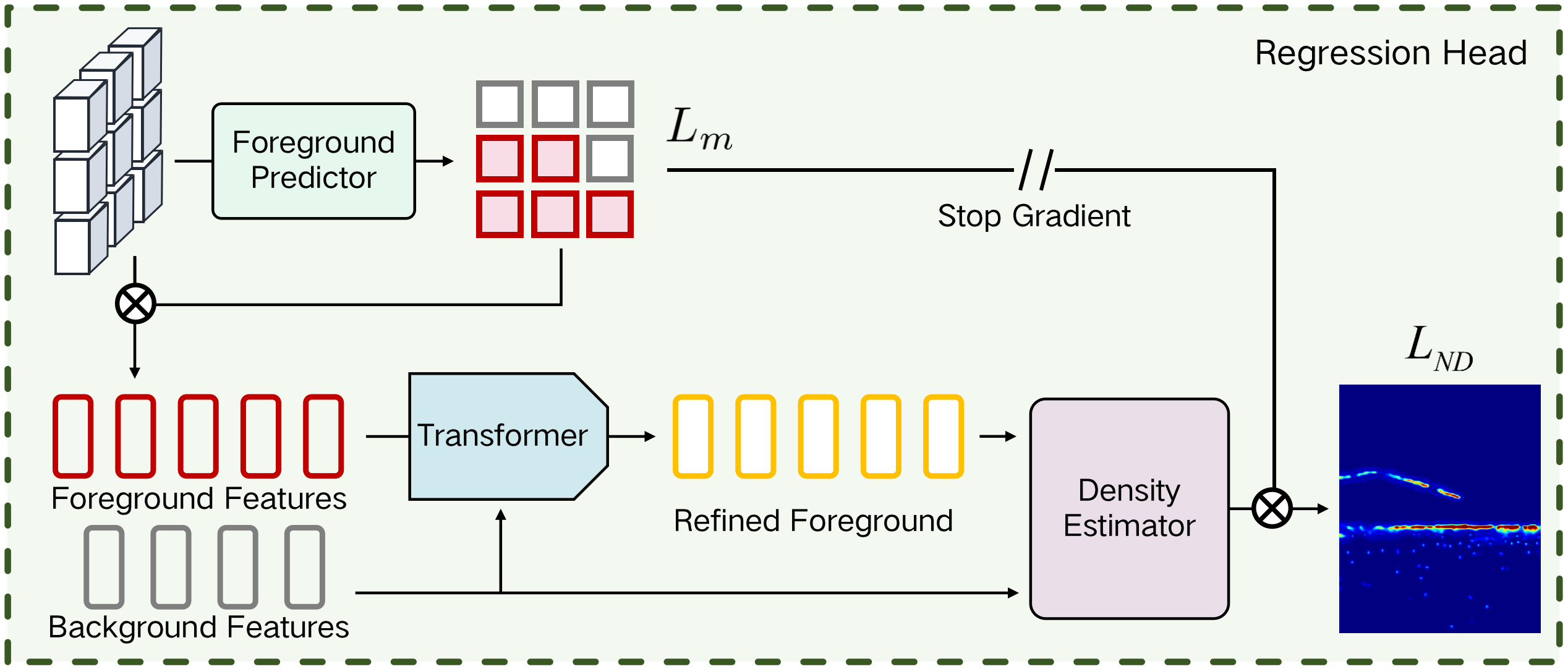}
\end{center}
\vspace{-3.5mm}
\caption{Details of the regression head. It consists of a foreground predictor to segment foreground and background, a transformer to refine foreground features and a density estimator to generate the density map. \XP{It is trained only using labeled data.}}
\label{fig:labeled}
\vspace{-5.5mm}
\end{figure}

The total framework of our counting model is shown in Figure~\ref{fig:labeled}. For an input image, we first use a CNN as the backbone to extract features. Then the regression head consists of three modules: a foreground predictor, a transformer and a density estimator, all of which are only trained by labeled data to remain 'pure'. 

The extracted features are first transmitted into the foreground predictor to get a division mask $M \in \{0,1\}^N$ where $N$ is the number of pixels. Then the foreground and background features can be divided by 
\begin{equation}\label{eq:foreback}
    \mathcal{E} \leftarrow {\rm sg}(M) \circ \mathcal{F}, \quad \mathcal{B} \leftarrow {\rm sg}(1-M) \circ \mathcal{F}.
\end{equation}
sg denotes stop gradients and $\circ$ denotes Hadamard product. The predicted foreground and background mask will be supervised by the generated \emph{binary} foreground mask $\hat{M}$ under MSE loss, which is defined by
\begin{equation}
    L_m = \sqrt{\sum (\hat{M}-M)^2}.
\end{equation}

\begin{figure}[t]
\begin{center}
    \includegraphics[width=0.45\textwidth]{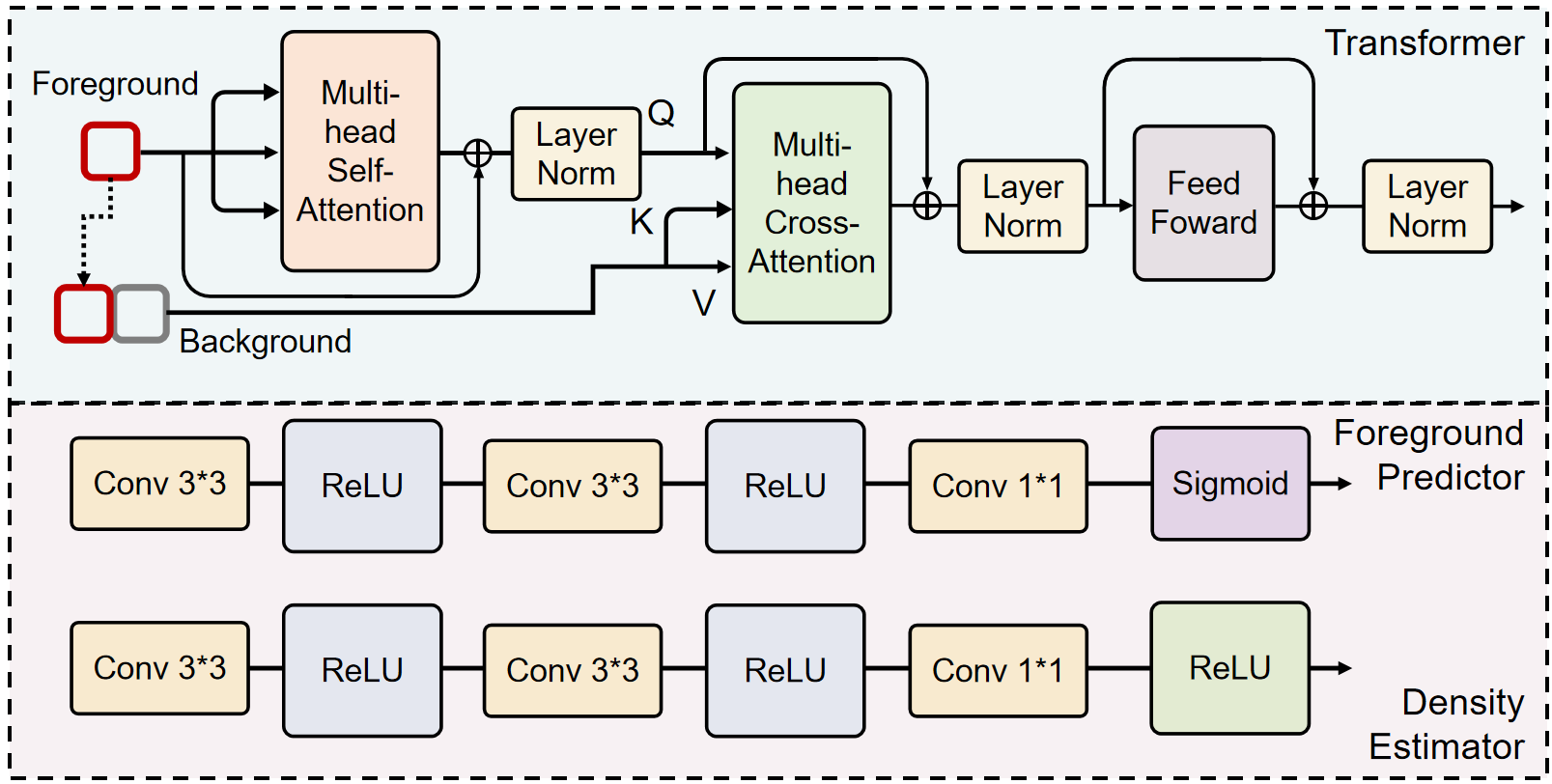}
\end{center}
\vspace{-4mm}
\caption{Detailed structures of the transformer, foreground predictor and density estimator.}
\label{fig:detail}
\vspace{-5.5mm}
\end{figure}

Next, we use the transformer structure to refine only the foreground features, which detail is shown in Figure~\ref{fig:detail}. It includes a self-attention module for foreground and a cross-attention between foreground and background features. With the help of these two modules, the foreground features can not only generate interactive attention with themselves, but also mine the relations with the background. The distance between the foreground and background features is further enlarged, and the differences within the foreground are also fully explored.

Finally, the background features and the refined foreground features are fed together into the density estimator to predict the final density map.

\subsection{Noise Depression Bayesian Loss}

%In order to maintain the accuracy 
\XP{To ensure the quality} of the regression head, only  labeled data are used to train it. However, there are \XP{still} location noises in the labels. Here, we propose a noise depression loss to minimize the negative influence to regression head by these noises.

We believe that when the gap between ground truth value and expected value the model predicted is too large, there may be noise in the label. Therefore, we propose to add a modulating factor to down-weigh these large gaps. Given the predicted density map D, the noise depression Bayesian loss is defined by
\begin{equation}\label{eq:focal}
\begin{aligned}
    & \varepsilon_j = |\delta(y_j) - \sum_{i}^N p_{ij} \cdot D_i|, \\
    &L_{ND} = \sum_{j=1}^M e^{-\beta{\rm sg}(\varepsilon_j)} \cdot \varepsilon_j. 
\end{aligned}
\end{equation}
${\rm sg(\cdot)}$ means stop gradients and $p_{ij}$ represents the contributed posterior probability of ground-truth $j$ for predicted pixel $i$. $\beta \geq 0$ is the depression parameter. When $\beta =0$, the loss will degenerate into normal Bayesian loss. Increasing the value of $\beta$, the loss will gradually ignore the penalty for large gap. we found $\beta = 1$ to work best in our experiments.
Then the total loss under labeled data turns into
\begin{equation}
    L_{label} = L_{ND} + \lambda_m L_m + \lambda_c L_C,
\end{equation}
while the loss under unlabeled data is
\begin{equation}
    L_{unlabel} =  \lambda_c L_C.
\end{equation}
And the total loss of the whole model is a combination of $L_{label}$ and $L_{unlabel}$ with an unlabeled parameter $\lambda_u$ is
\begin{equation}
    L = L_{label} + \lambda_u L_{unlabel}.
\end{equation}
\XP{The training process of DACount is summarised in Algorithm 1.}
%\XP{The training process of the proposed DACount is summarised in Algorithm 1.}
%~\ref{algo:dacount}. 

\begin{algorithm}

\caption{\XP{DACount Learning}}
%Procedure for the training of proposed model $R$ and the density agent $\mathcal{A}$.
\LinesNumbered 
\KwIn{Sampled labeled images $\mathcal{X}_l \in \mathcal{L}$ and unlabeled images $\mathcal{X}_u \in \mathcal{U}$.}
\XP{\KwOut{The counting model $R$ which consists of feature backbone $g$ and regression head $h$, $R = h \circ g$. The density agent $\mathcal{A}$.}}
Initialize $L_\mathcal{A} \leftarrow 0$, $L \leftarrow 0$\;
\For{$x \in \mathcal{X}_l \cup \mathcal{X}_u$}{
    Get features $\mathcal{F} \leftarrow g(x)$\;
    Get foreground mask and predicted density map $M, D \leftarrow h(\mathcal{F})$\;
  \If{$x \in \mathcal{X}_l$}{
      Generate ground-truth foreground mask $\hat{M}$ and ground-truth density map $\hat{D}$\;
    $L \leftarrow L + L_{ND}$ based on Eq.~\ref{eq:focal}\;
    $L \leftarrow L + \lambda_m L_m$ based on MSE loss between $M$ and $\hat{M}$\;
    Update regression head $h$ minimizing $L$\;}
    % Get foreground and background features based on Eq.~\ref{eq:foreback} replacing $M$ by $\hat{M}$\;
    % Divide $\mathcal{E}$ by $\hat{D}$\;
    Get foreground and background features based on Eq.~\ref{eq:foreback} by $M$ or $\hat{M}$\;
    Divide $\mathcal{E}$ by $D$ or $\hat{D}$\;
%   }{
%       Get foreground and background features based on Eq.~\ref{eq:foreback}\;
%     Divide $\mathcal{E}$ by $D$\;
%   }
  Calculate $L_\mathcal{E}$ and $L_\mathcal{B}$ based on Eq.~\ref{eq:foreground} and Eq.~\ref{eq:background}\;
%   Calculate $L_\mathcal{B}$ based on Eq.~\ref{eq:background}\;
  Calculate $L_C$ based on Eq.~\ref{eq:contrastive} and Eq.~\ref{eq:contrastive2}\;
  $L \leftarrow L + \lambda_{c}L_C$\;
  $L_{\mathcal{A}} \leftarrow L_{\mathcal{A}} + L_\mathcal{E} + L_\mathcal{B}$\;
  \If{$x \in \mathcal{X}_u$}{Adjust the loss weight by $L \leftarrow \lambda_u L$, $L_\mathcal{A} \leftarrow \lambda_u L_{\mathcal{A}}$\;}
}

Update $\mathcal{A}$ minimizing $L_{\mathcal{A}}$\;
Update feature backbone $g$ minimizing $L$\;
\XP{\textbf{Return}} {the model $R=h \circ g$, the density agent $\mathcal{A}$.}
\label{algo:dacount}
\end{algorithm}
%-------------------------------------------------------------------------

% \def\arraystretch{1.1}
\renewcommand{\tabcolsep}{12 pt}{
\begin{table*}[t!]
\small
	\begin{center}
		\begin{tabular}{c|c|cc|cc|cc|cc}
			\toprule[1.5pt]
			\multicolumn{1}{c}{\multirow{2}*{Methods}} & \multicolumn{1}{c}{Labeled} &  \multicolumn{2}{c}{UCF-QNRF} &  \multicolumn{2}{c}{JHU++} &
			\multicolumn{2}{c}{ShanghaiTech A} &
			\multicolumn{2}{c}{ShanghaiTech B} \\
			
			\multicolumn{1}{c}{} & \multicolumn{1}{c}{Percentage} & \multicolumn{1}{c}{MAE} &
			\multicolumn{1}{c}{MSE} & \multicolumn{1}{c}{MAE} &
			\multicolumn{1}{c}{MSE} &  \multicolumn{1}{c}{MAE} &
			\multicolumn{1}{c}{MSE} &  \multicolumn{1}{c}{MAE} &
			\multicolumn{1}{c}{MSE}\\
			\hline
			\hline
			MT~\cite{tarvainen2017mean} & 5\%  & 172.4 & 284.9 & 101.5 & 363.5 & 104.7 & 156.9 & 19.3 & 33.2\\
			
			L2R~\cite{liu2018leveraging}  & 5\% & 160.1 & 272.3 & 101.4 & 338.8 & 103.0 & 155.4 & 20.3 & 27.6 \\
			GP~\cite{sindagi2020learning}  & 5\% & 160.0 & 275.0 & - & -  & 102.0 & 172.0 & 15.7 & 27.9 \\
			DACount (Ours)  & 5\% & \textbf{120.2} & \textbf{209.3} & \textbf{82.2} & \textbf{294.9}  & \textbf{85.4} & \textbf{134.5} & \textbf{12.6} & \textbf{22.8}  \\
			\midrule
			MT~\cite{tarvainen2017mean}  & 10\% & 145.5 & 250.3 & 90.2 & 319.3  & 94.5 & 156.1 & 15.6 & 24.5 \\
		    L2R~\cite{liu2018leveraging} & 10\% & 148.9 & 249.8 & 87.5 & 315.3 & 90.3 & 153.5 & 15.6 & 24.4 \\
		    IRAST~\cite{liu2020semi} & 10\% & - & - & - & -  & 86.9 & 148.9 & 14.7 & 22.9\\
		    IRAST+SPN~\cite{liu2020semi} & 10\%  & - & - & - & - & 83.9 & 140.1 & - & - \\
		    DACount (Ours)  & 10\% & \textbf{109.0} & \textbf{187.2} & \textbf{75.9} & \textbf{282.3}  & \textbf{74.9} & \textbf{115.5} & \textbf{11.1} & \textbf{19.1} \\
		    \midrule
			MT~\cite{tarvainen2017mean}  & 40\% & 147.2 & 249.6 & 121.5 & 388.9 & 88.2 & 151.1 & 15.9 & 25.7 \\
		    L2R~\cite{liu2018leveraging} & 40\% & 145.1 & 256.1 & 123.6 & 376.1 & 86.5 & 148.2 & 16.8 & 25.1 \\
		    GP~\cite{sindagi2020learning}  & 40\% & 136.0 & - & - & - & 89.0 & - & - & - \\
		    IRAST~\cite{liu2020semi} & 40\% & 138.9  & - & - & - & - & - & - & - \\
		    SUA~\cite{meng2021spatial} & 40\% & 130.3 & 226.3 & 80.7 & 290.8 & 68.5 & 121.9 & 14.1 & 20.6 \\
		    DACount (Ours)  & 40\% & \textbf{91.1} & \textbf{153.4} & \textbf{65.1} & \textbf{260.0}   & \textbf{67.5} & \textbf{110.7} & \textbf{9.6} & \textbf{14.6}\\
			\bottomrule[1.5pt]
		\end{tabular}
	\end{center}

\caption{Comparisons with the state of the arts semi-supervised counting methods on ShanghaiTech A, ShanghaiTech B, UCF-QNRF, and JHU-Crowd++. The best performance is shown in \textbf{bold}. \XP{The results of other methods under the $40\%$ labeled setting} are referred to \cite{meng2021spatial} \XP{and all other results are from the original papers.}}
\vspace{-4mm}
\label{tab:performance}
\end{table*}}
%Results with $40\%$ labeled of other methods are referred to \cite{meng2021spatial}.

\section{Experiments}
\subsection{Implementation Details: }
{\flushleft\textbf{Network Details: }}
VGG-19 is pre-trained on ImageNet and adopted as our CNN backbone. The number of density agents is set as $N_a = 24$. We use Adam algorithm~\cite{kingma2014adam} to optimize the model and the agent. Both set the learning rate as $10^{-6}$, specifically, $\gamma=10^{-6}$. For the loss parameters, we set $\lambda_m = 0.1$, $\lambda_c = 0.01$, and $\lambda_u = 0.1$. The depression parameter $\beta = 1$. The temperature $\tau = 0.1$.

{\flushleft\textbf{Training Details: }}
We adopt random scaling of [0.7, 1.3] and horizontal flipping for each training image, \XP{and use random crop with a size of 512 × 512. The only exception is to use the size of 256 × 256 for extremely small images on ShanghaiTech A.}
%And the random crop with a size of 512 × 512 is implemented. As some images in ShanghaiTech A contain smaller resolution, the crop size for this dataset reduces to 256 × 256. 
We limit the shorter side of each image within 2048 pixels in all datasets.

% {\flushleft\textbf{Evaluation Metrics: }}We evaluate the methods on testing set by two commonly used metrics: Mean Absolute Error (MAE) and Mean Squared Error (MSE). They are defined as follows:
% \begin{equation}
% \begin{aligned}
% MAE=\frac{1}{N_t}\ \sum_{i=1}^{N_t}\big|C_i^{gt}-C_i\big|,\\ MSE=\sqrt{\frac{1}{N_t}\ \sum_{i=1}^{N_t}(C_i^{gt}-C_i)^2)}.
% \end{aligned}
% \end{equation}
% where $N_t$ is the number of sample images. $C_i^{gt}$ and $C_i$ are ground truth and estimated count of $i_{th}$ image respectively. MAE mainly measures the accuracy of methods and MSE mainly measures the robustness. The lower of both represents the better performance~\cite{zhang2016single}. 

\subsection{Datasets}
We conduct extensive experiments on four crowd counting benchmarks to verify the effectiveness of our proposed method.
%We have chosen four largest datasets, which are described as follows:

{\flushleft\textbf{ShanghaiTech A}~\cite{zhang2016single}} contains 482 crowd images with 244,167 annotated points. %These images are randomly chosen from the Internet. 
%The number of annotations in an image ranges from 33 to 3139.
The training set has 300 images, and the testing set has the remaining 182 images. 

{\flushleft\textbf{ShanghaiTech B}~\cite{zhang2016single}} contains 716 crowd images, which are taken in the crowded street of Shanghai. %The number of annotations in an image ranges from 9 to 578. 
The training set has 316 images, and the testing set has the remaining 400 images. 
    
{\flushleft\textbf{UCF-QNRF}~\cite{idrees2018composition}} includes 1535 high-resolution images with 1.25 million annotated points. 
%It is a congested crowd dataset, which images are crawled from Flickr, Web Search, and Hajj footage. 
There are 1,201 images in the training set and 334 images in the testing set.
%, with the average number of persons per image 815.
%, and the minimum and maximum numbers are 49 and 12,865, respectively.
    
{\flushleft\textbf{JHU-Crowd++}~\cite{sindagi2020jhu}} includes 4,372 images with 1.51 million annotated points. There are 2272 images used for training, 500 images for validation, and the rest 1600 images used for testing. 
%It is a large-scale dataset, which images are collected from several sources on the Internet using different keywords and specifically chosen under diverse conditions and various geographical locations.
    
% {\flushleft\textbf{NWPU-CROWD}~\cite{wang2020nwpu}} contains 5,109 images with 2.13 million annotated points. 3,109 images are used for training, 500 images are used for validation, and the remaining 1,500 images are used for testing. Compared with other datasets, it has significant scale variations, where the number of annotated people in a single image ranges from 0 to 20,033 and contains various illumination scenes. It also includes 351 negative samples of nobody or textures similar to congested crowd scenes, making the dataset much more challenging.
\renewcommand{\tabcolsep}{5 pt}{
\begin{figure*}[t!]
	\begin{center}
		\begin{tabular}{cccc}
			
			\includegraphics[height=0.14\linewidth]{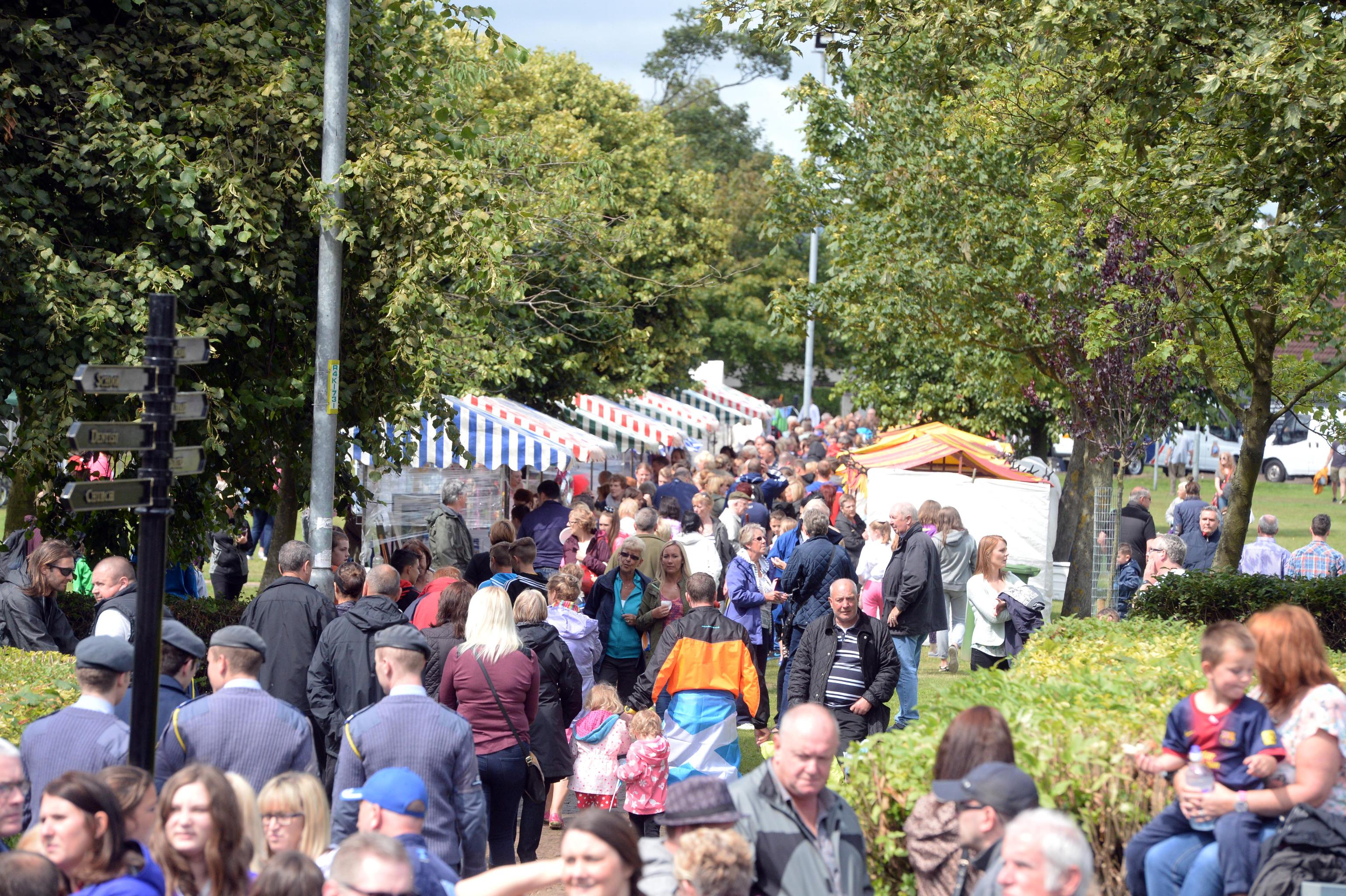}  &
			\includegraphics[height=0.14\linewidth]{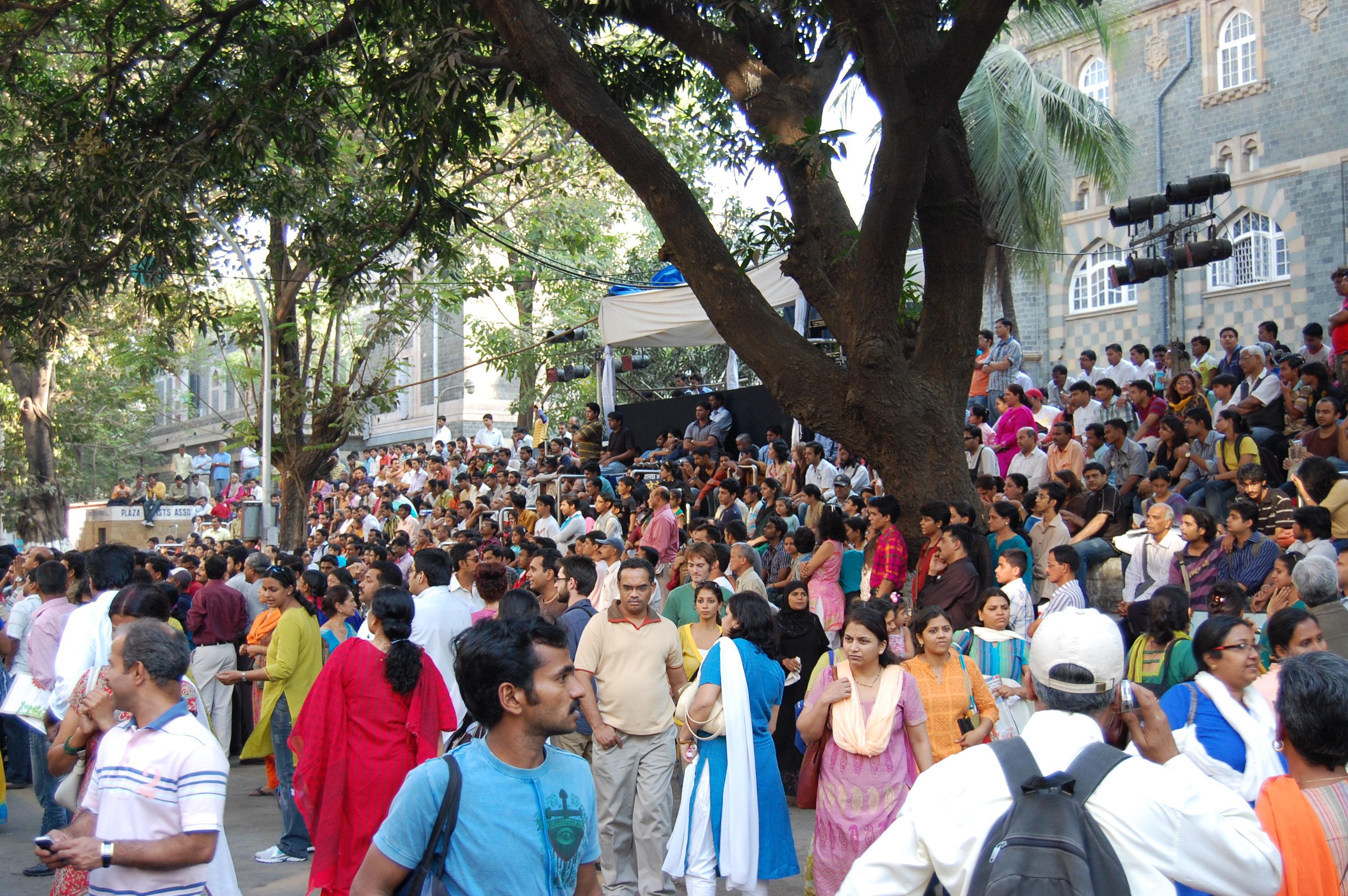} &
			\includegraphics[height=0.14\linewidth]{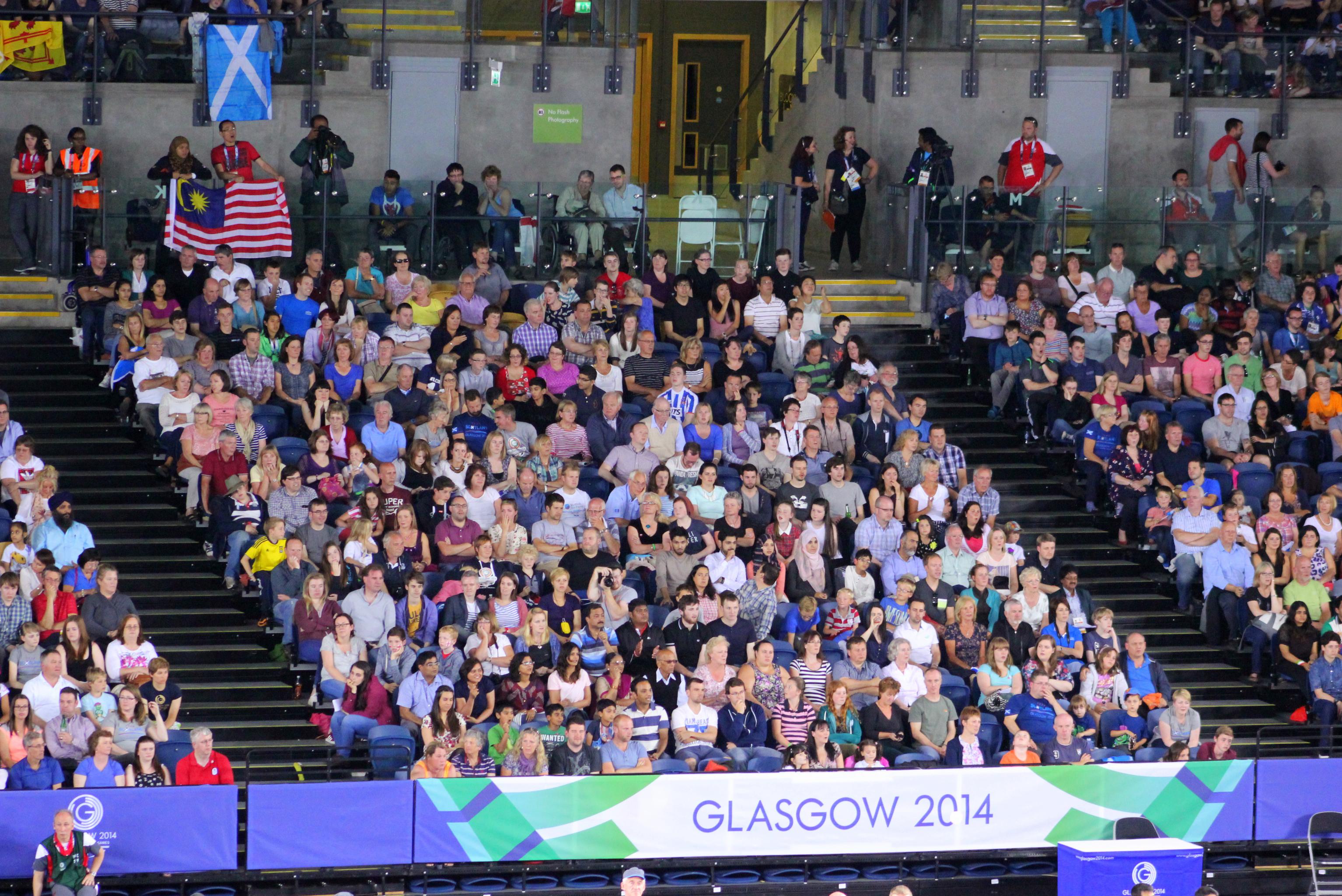}  &
			\includegraphics[height=0.14\linewidth]{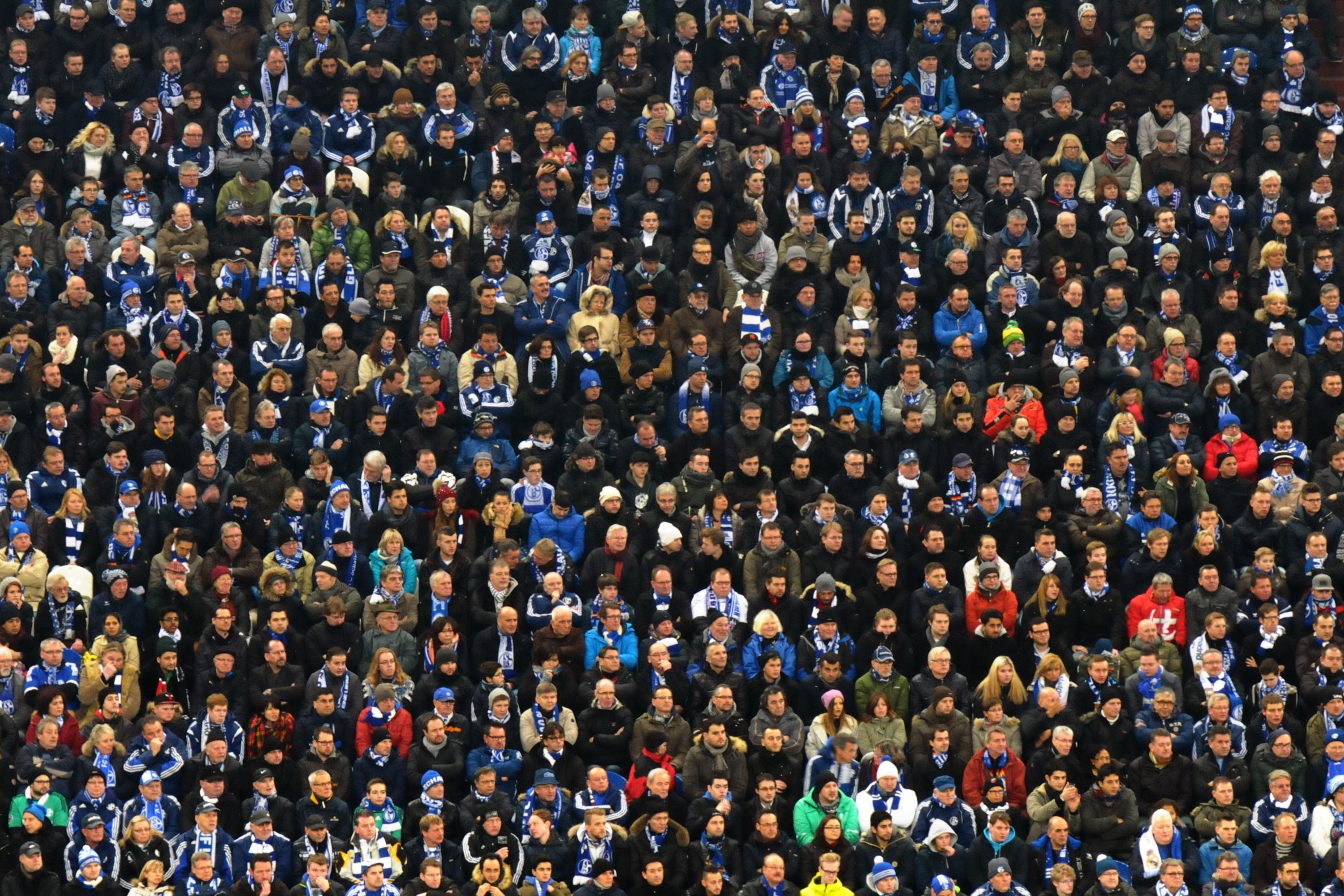} \\
			
			\footnotesize{GT: 137} & \footnotesize{GT: 337} & \footnotesize{GT: 349} & \footnotesize{GT: 558} \\
			
			\includegraphics[height=0.14\linewidth]{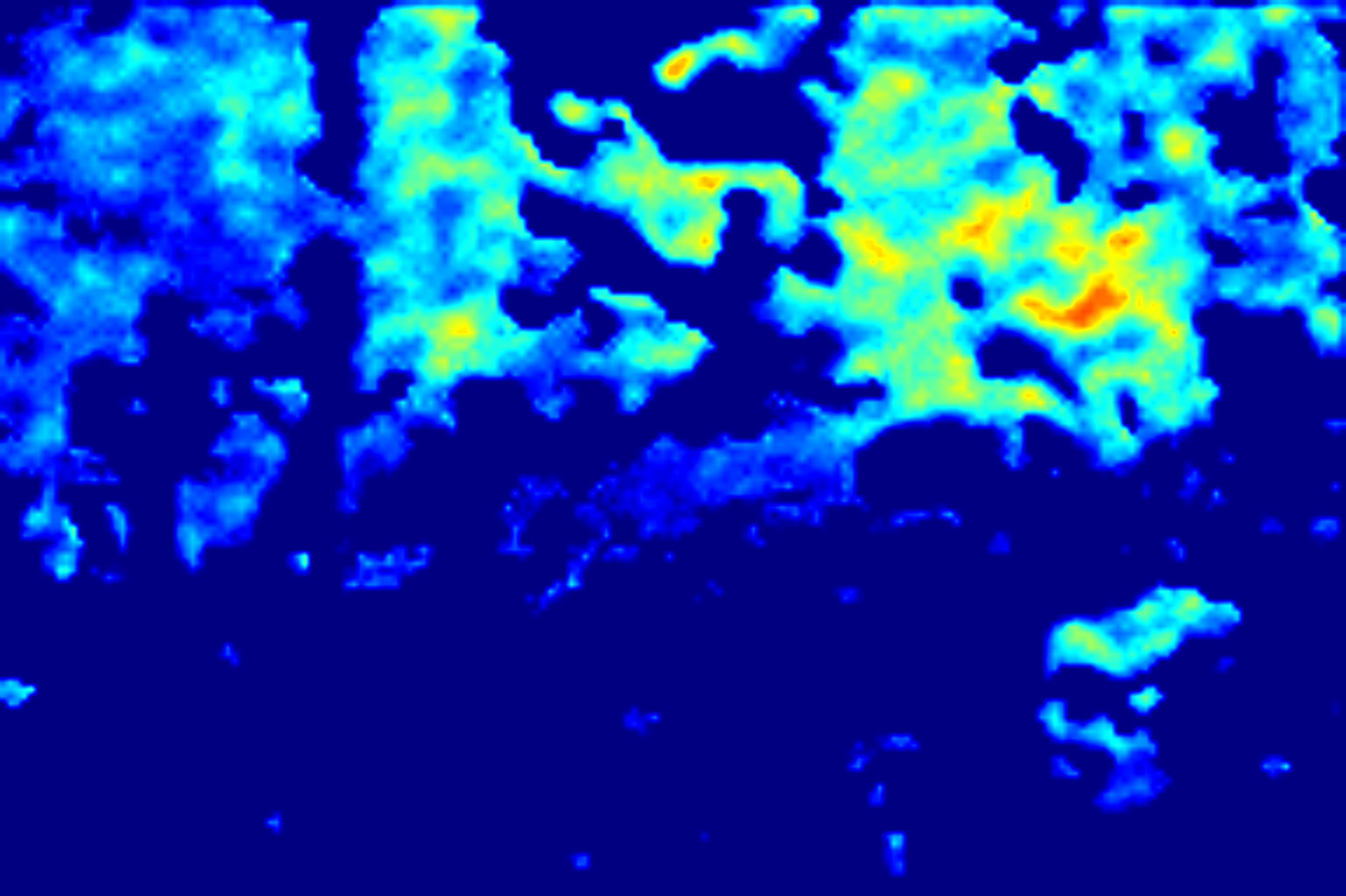}  &
			\includegraphics[height=0.14\linewidth]{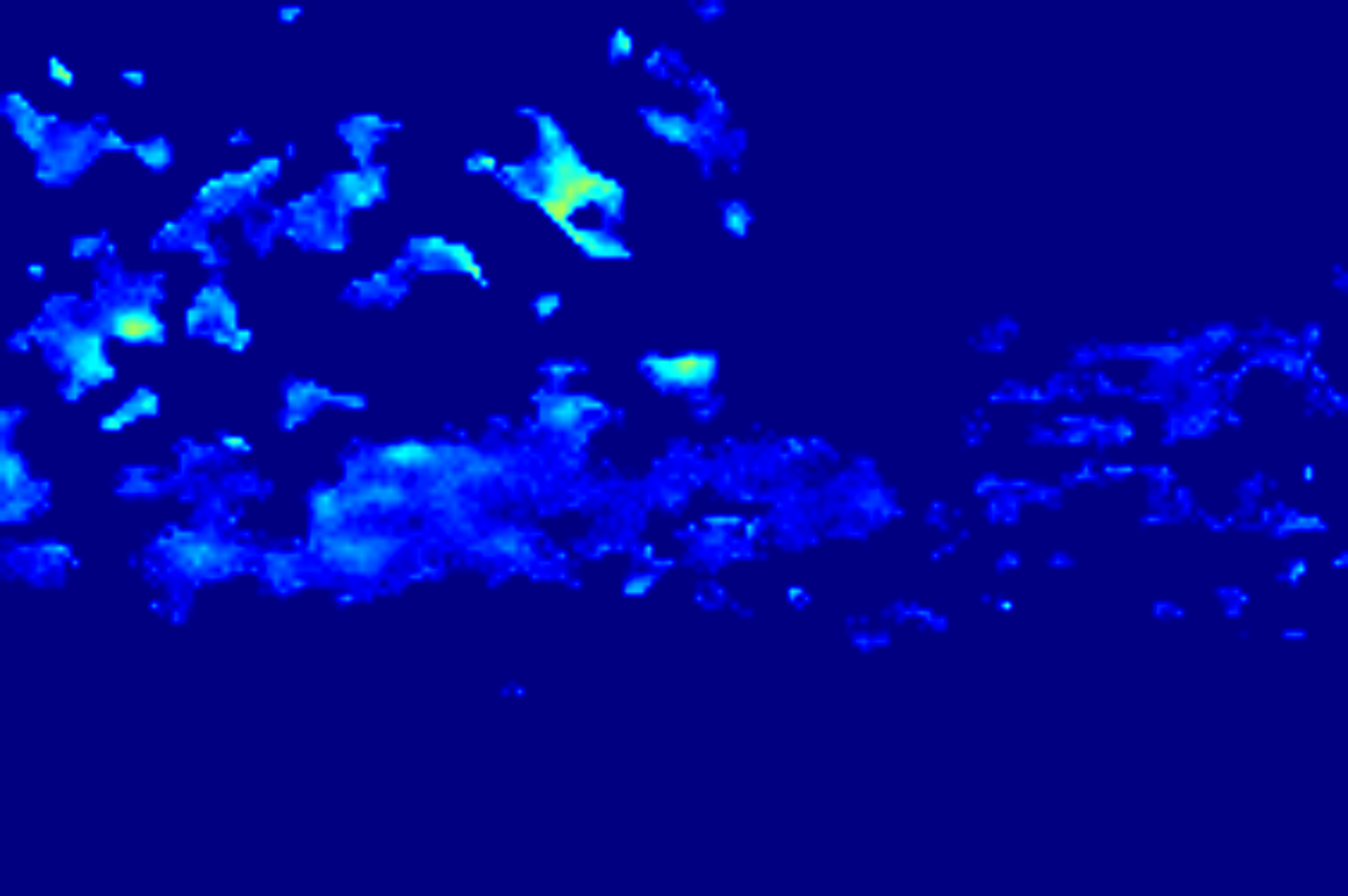} &
			\includegraphics[height=0.14\linewidth]{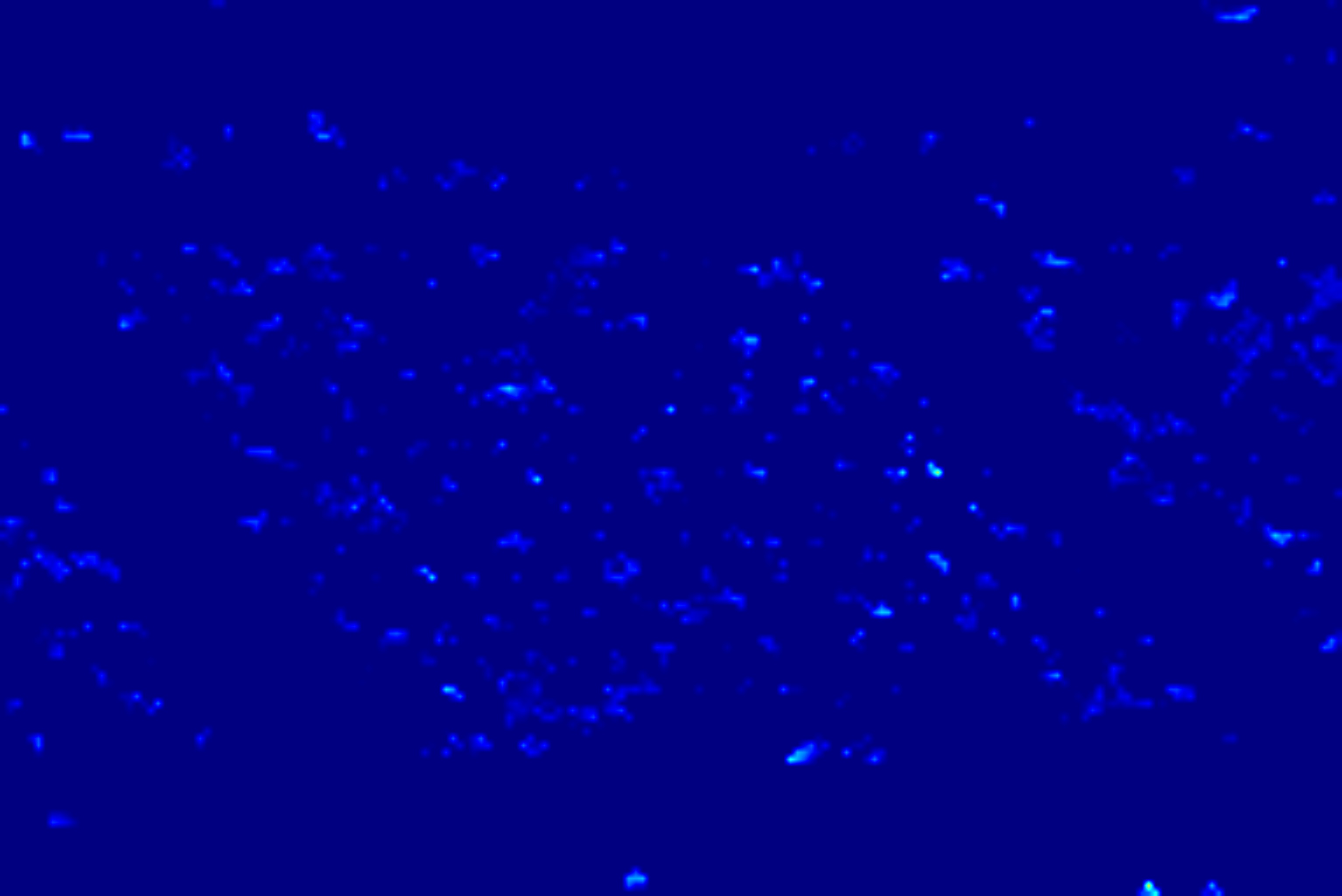}  &
			\includegraphics[height=0.14\linewidth]{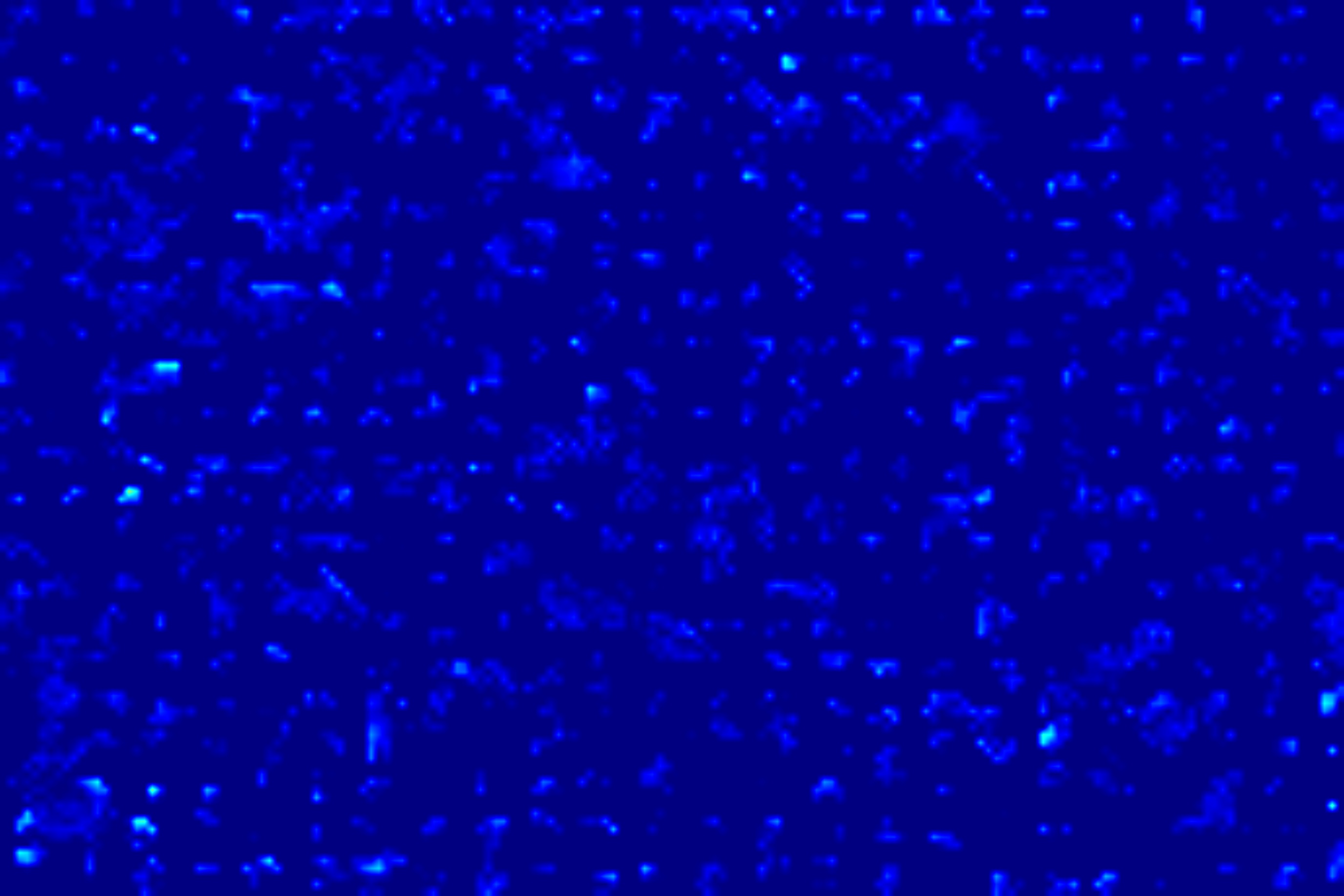} \\
			
			\footnotesize{Labeled Only: 2963} & \footnotesize{Labeled Only: 617} & \footnotesize{Labeled Only: 131} & \footnotesize{Labeled Only: 345} \\
			\includegraphics[height=0.14\linewidth]{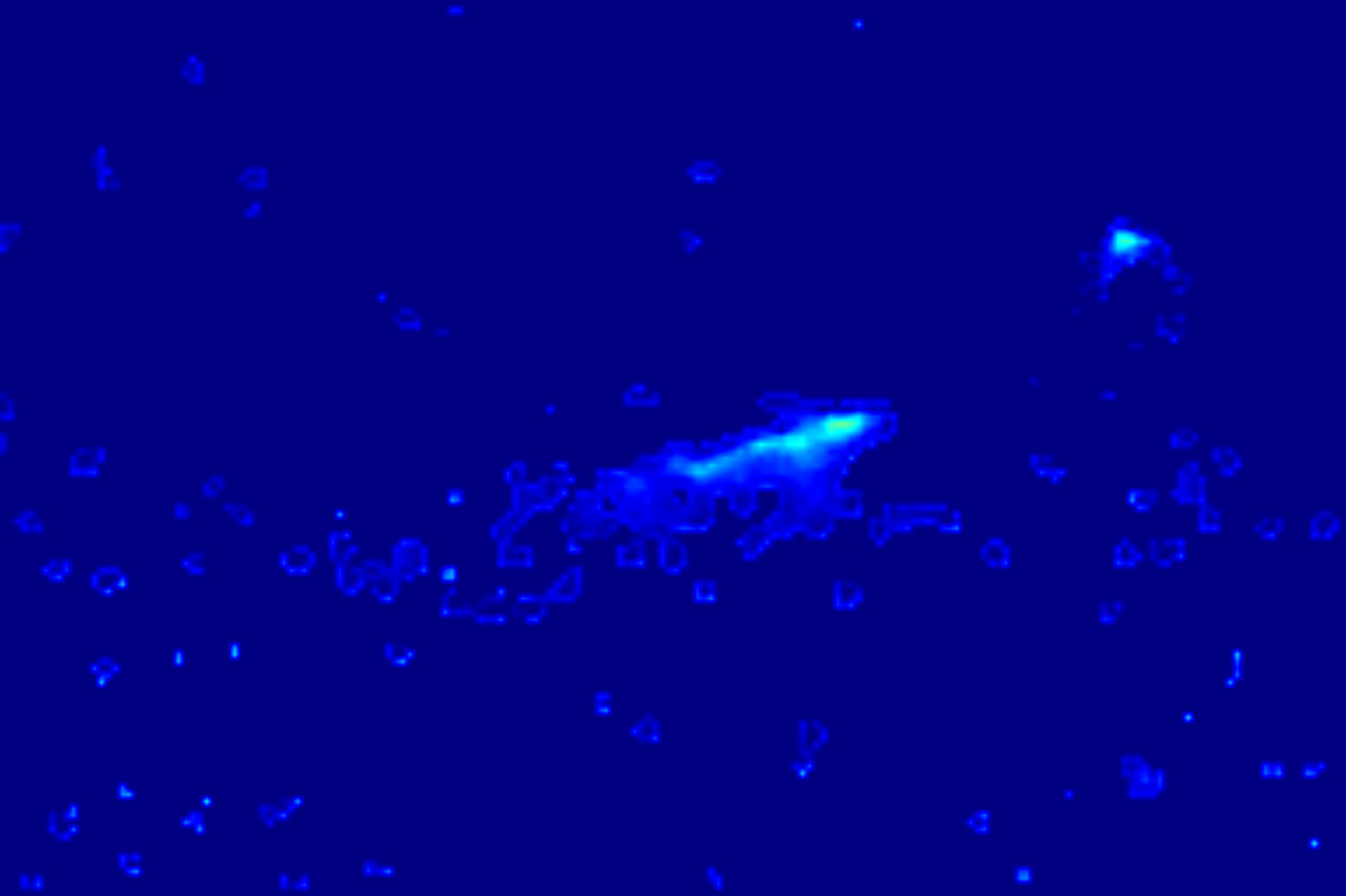}  &
			\includegraphics[height=0.14\linewidth]{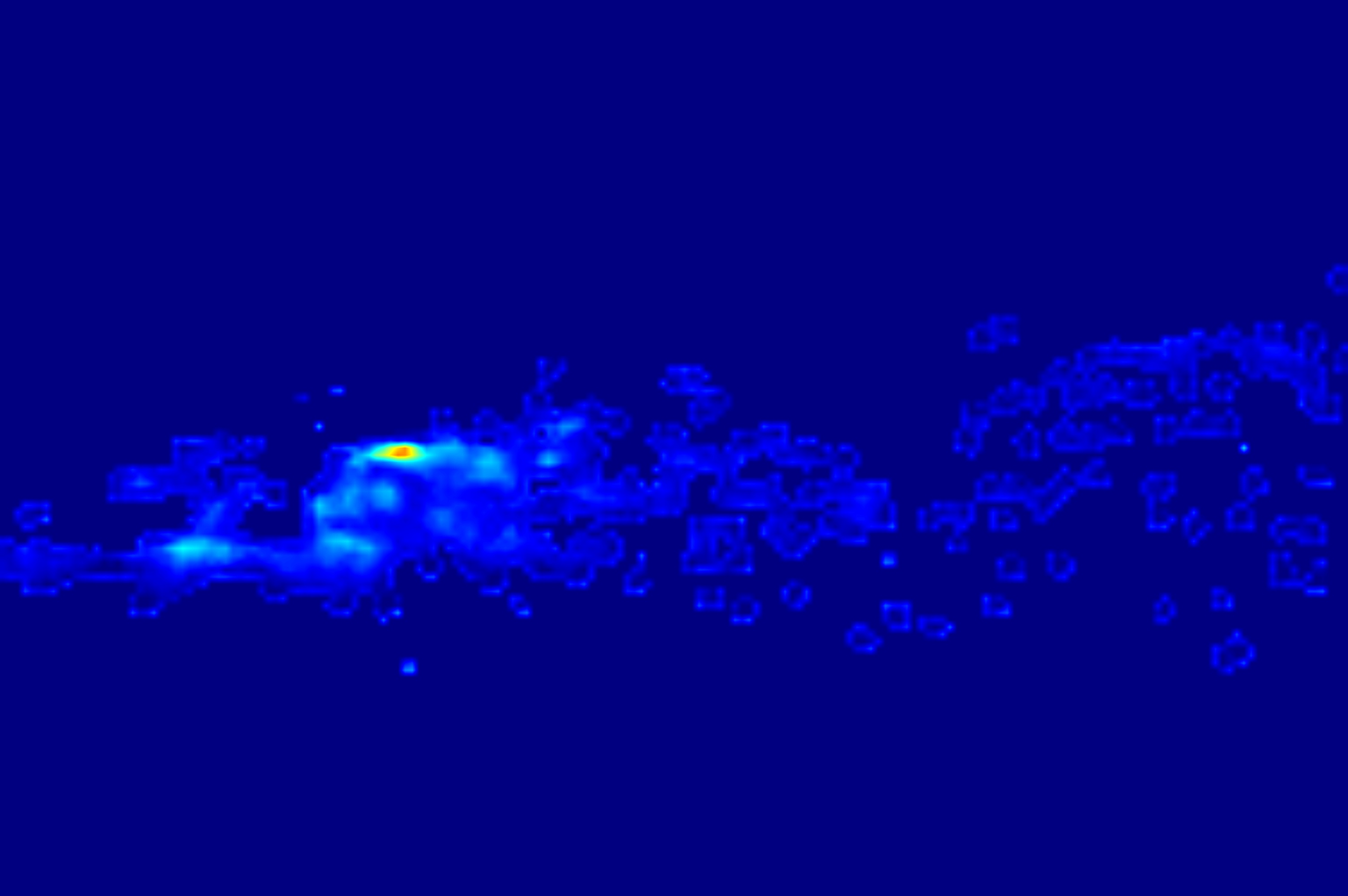} &
			\includegraphics[height=0.14\linewidth]{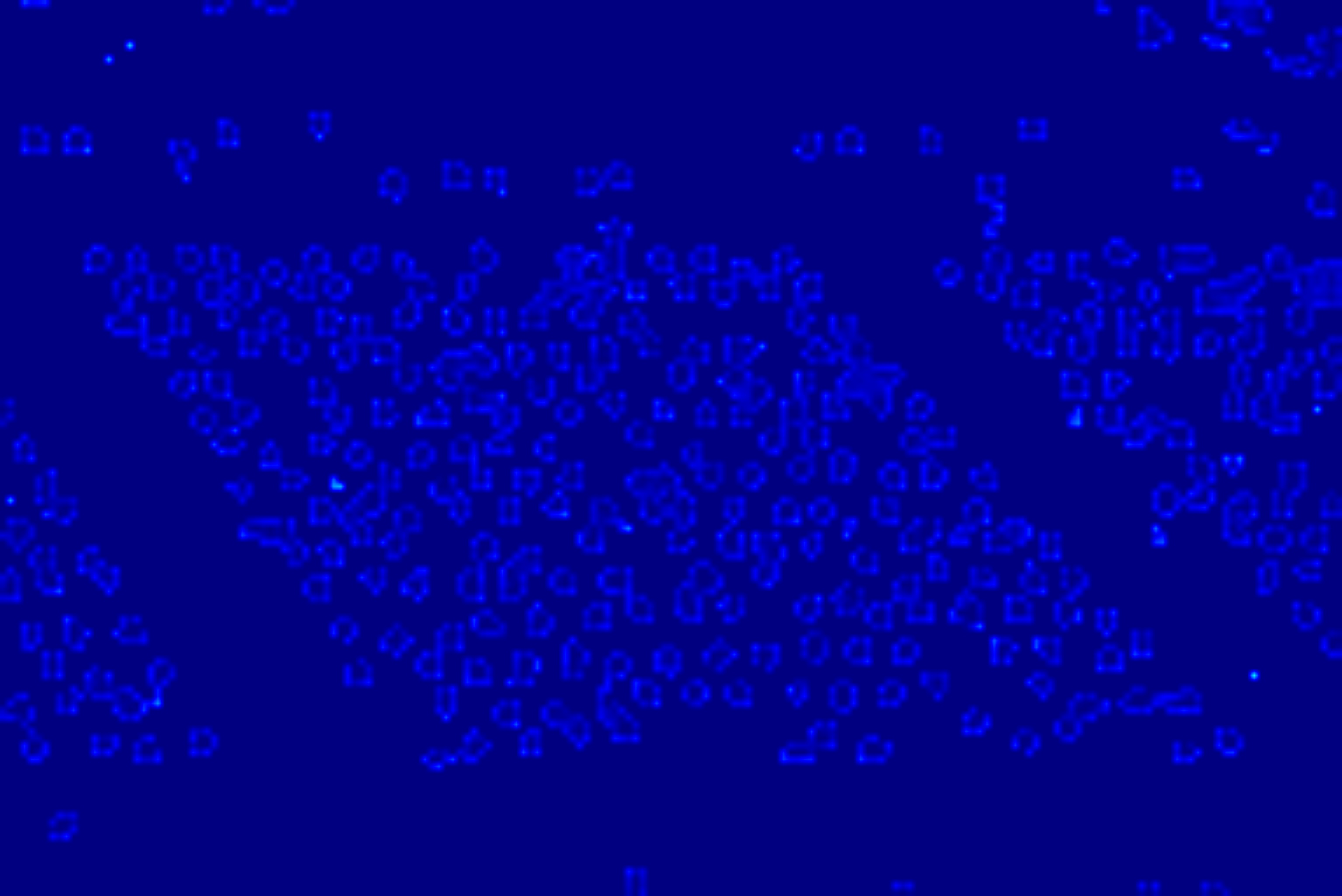}  &
			\includegraphics[height=0.14\linewidth]{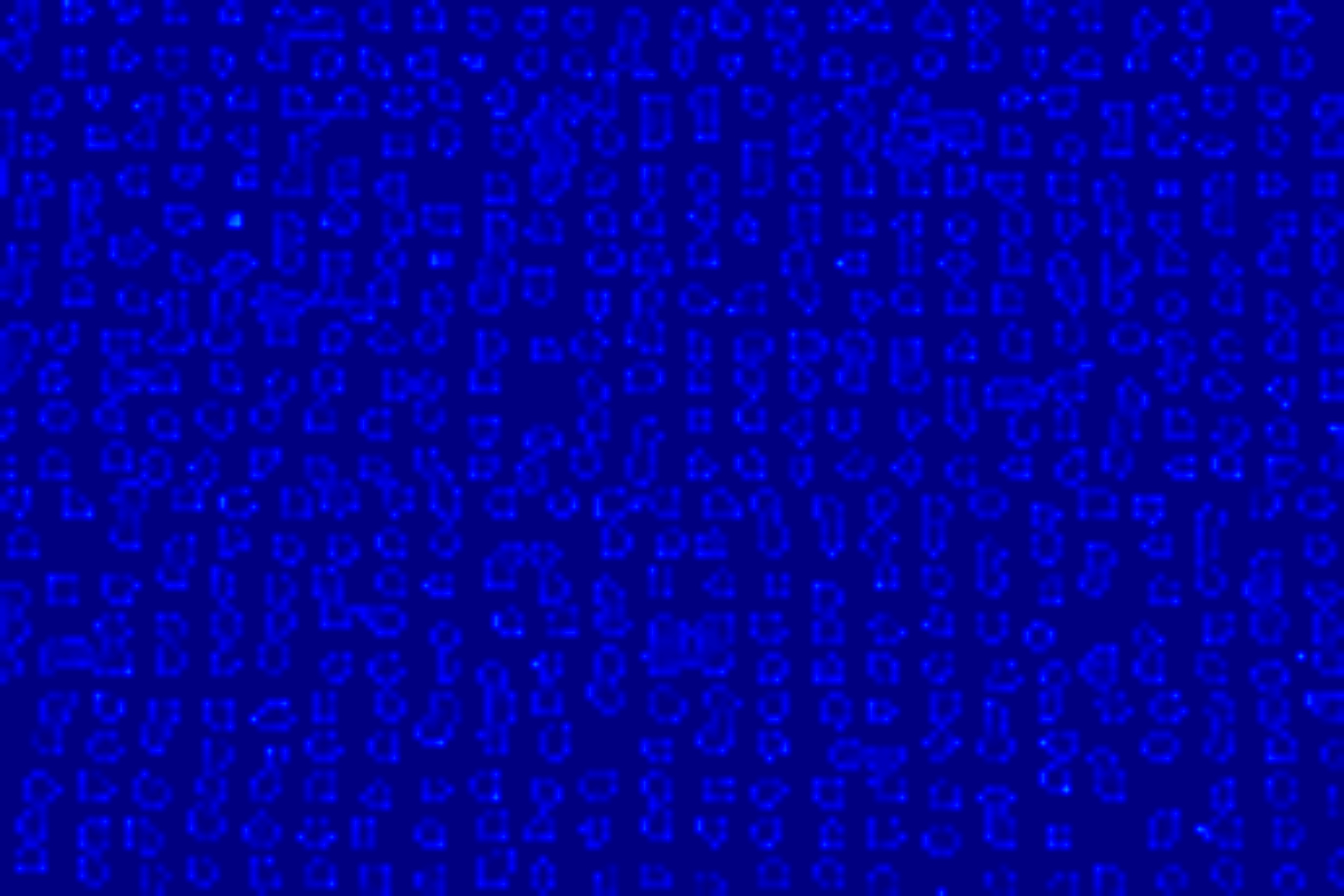} \\
			
			\footnotesize{DACount: 174} & \footnotesize{DACount: 323} & \footnotesize{DACount: 345} & \footnotesize{DACount: 534} \\

		\end{tabular}
		\vspace{-4mm}
		\caption{Visualizations of predicted densities on unlabeled training images. The first row: input images. The second row: predicted density maps trained only by labeled data. The third {row}: predicted density maps trained by both labeled and unlabeled data. Both models are trained on UCF-QNRF with a labeled ratio of $5\%$. {When the model is trained only with labeled data, serious false alarms in the background (the first two examples) and apparent underestimation of the foreground density (the last two examples) are observed in the second row. In contrast, our agency-guided semi-supervised counting scheme can well exploit rich supervised signal from unlabeled data and thus produce density maps of much higher quality in the third row.}}
		% The first two columns indicate that the density maps trained only with labeled data will incorrectly learn the background. The other two columns show labeled-only-densities are relatively fewer predicted.
		\label{fig:viz}
	\end{center}
\vspace{-4mm}
\end{figure*}}

\subsection{Experimental Results}
We evaluate DACount and compare it with state-of-the-art semi-supervised methods, including mean-teacher (MT)~\cite{tarvainen2017mean}, Learning to Rank (L2R)~\cite{liu2018leveraging}, Gaussian Process (GP)~\cite{sindagi2020learning}, IRAST~\cite{liu2020semi}, IRAST+SPN (Scale Pyramid Network)~\cite{liu2020semi}, and SUA~\cite{meng2021spatial}.%The result is shown in Table~\ref{tab:performance}. \XP{The results of other methods under the $40\%$ labeled setting} are referred to \cite{meng2021spatial} \XP{and all other results are from the original papers.}

\XP{The results are shown in Table~\ref{tab:performance}. It can been easily observed that} DACount outperforms \XP{other} state-of-the-art semi-supervised methods by a significant accuracy improvement on all datasets, \XP{regardless of the settings of the ratio of labeled data. Under the (relatively) easiest setting (i.e., the setting of 40\% labeled data), our method reduces the MAE by 39.2, 15.6, 1.0, and 4.5 points on databases QNRF, JHU++, SHanghaiTech A and B respectively, compared to the second best method SUA. The improvement becomes even more apparent when there is less labeled data available. For example,} 
%Our method achieves pleasing results starting from the least labeled ratio of $5\%$, and maintains the great performance consistently at $10\%$ and $40\%$. Especially, 
on UCF-QNRF dataset, DACount reduces all the previous methods by \XP{at least 39.8 and 62.6 in terms of} MAE and MSE, respectively. These excellent results demonstrate the effectiveness of our method in semi-supervised crowd counting.

%DACount outperforms \XP{other} state-of-the-art semi-supervised methods by a significant accuracy improvement on all datasets, especially when the labeled data is limited. Our method achieves pleasing results starting from the least labeled ratio of $5\%$, and maintains the great performance consistently at $10\%$ and $40\%$. Especially, on UCF-QNRF dataset, DACount reduces all the previous methods over by $24.6\%$ and $18.7\%$ for MAE and MSE, respectively. These excellent results demonstrate the effectiveness of our method in semi-supervised crowd counting.

\subsection{Ablation study}

%As aforementioned, our proposed DACount is composed of the following components: the learnable density agent, contrastive loss, and noise depression loss.
We conduct ablation experiments to verify the effectiveness of the three component in DACount, namely the learnable density agent, contrastive loss, and noise depression loss, as shown in Table~\ref{tab:ablation}.

\renewcommand{\tabcolsep}{6 pt}{
\begin{table}[t]
\small
\begin{center}
\begin{tabular}{lcc}
  \toprule[1pt]
  Components & MAE & MSE\\
  \midrule
  Baseline (BL) & 151.4 & 263.9 \\
  $+$ Transformer & 141.3 & 241.2 \\
  $+$ Momentum Density Agent & 130.6 & 218.3\\
  $+$ Learnable Density Agent & 128.4 & 215.9 \\
  $+$ Contrastive Learning & 124.7 & 214.6 \\
  $+$ Uncertainty Aware Contrastive & 122.0 & 212.9 \\
  $+$ Noise Depression Loss & \textbf{120.2} & \textbf{209.3} \\
  \toprule[1pt]
\end{tabular}
\caption{Ablation study on UCF-QNRF (labeled ratio $5\%$).}
% \caption{Ablation study. Experiments are conducted on UCF-QNRF with a labeled ratio $5\%$. The model achieves its best when combining all proposed components.}
\vspace{-8mm}
\label{tab:ablation}
\end{center}
\end{table}}
%for different components

All experiments are conducted on UCF-QNRF with a labeled ratio of $5\%$. We start with the baseline of BL~\cite{ma2019bayesian}, which only uses labeled data and vanilla CNN to extract features. Then, we adopt transformer module with foreground and background separated, which detail can be referred to Section~\ref{sec:structure}, and the performance improves by a large margin even still without unlabeled data. We believe that this improvement is due to the global attention of foreground on the one hand, and the cross attention with background on the other hand. 

Next, we adopt the proposed density agency to study the help of this supervision on unlabeled data. With the agency, we can take full advantage of the potential of unlabeled data. The learnable structure builds a bridge between labeled and unlabeled data, exploiting the correlations among different images. The error reduces $12.9$ and $25.3$ for MAE and MSE, respectively. Then by transferring density information to features and widening the differences between foreground, the contrastive learning further improves the counting accuracy. In addition, we also take into account the uncertainty of the predicted density map, weighting the positive and negative counterparts, and have a pleasing performance. Finally, the noise depression Bayesian loss down-weights the supervision of noisy instances. The best performance is achieved when combining all above components, improving $31.2$ and $54.6$ for MAE and MSE compared to the baseline.

\subsection{Discussions}
{\flushleft\textbf{The impact of unlabeled data. }}We conduct experiments to examine the impact of unlabeled data. Specifically, excluding unlabeled data can be equal to only using the labeled loss $L_{label}$. And the combination of $L_{label}$ and $L_{unlabel}$ forms the proposed DACount. The comparison result is shown in Table~\ref{tab:unlabeled}.

Compared to using labeled data only, DACount which leverages on unlabeled data outperforms consistently with all labeled percentages. Specifically, it keeps a over $7.4\%$ improvement with the help of unlabeled data. As the results shown, when the labeled percentage is smaller, the improvement will be larger. This is reasonable since the model can utilize more unlabeled data. This quantitative experiment proves that our model can make full use of unlabeled data to help improve the performance.

\renewcommand{\tabcolsep}{6 pt}{
\begin{table}[t]
\small
\begin{center}
\begin{tabular}{cccc}
  \toprule[1pt]
  Loss & Labeled Percentage & MAE & MSE\\
  \midrule
  $L_{label}$ & $5\%$ & 138.3 & 241.1 \\
  $L_{label} + \lambda_u L_{unlabel}$ & $5\%$ & 120.2 & 209.3 \\
  \midrule
  $L_{label}$ & $10\%$ & 129.4 & 229.7 \\
  $L_{label} + \lambda_u L_{unlabel}$ & $10\%$ & 109.0 & 187.2 \\
  \midrule
  $L_{label}$ & $40\%$ & 98.4 & 182.7 \\
  $L_{label} + \lambda_u L_{unlabel}$ & $40\%$ & 91.1 & 153.4 \\
  \toprule[1pt]
\end{tabular}
\caption{The impact of {using} unlabeled data on UCF-QNRF.} \label{tab:unlabeled}
\vspace{-9mm}
%\caption{The impact of \XP{using} unlabeled data. Experiments are conducted on UCF-QNRF. Our method can take advantage of unlabeled data to achieve high accuracy. With the help of unlabeled data, the method outperforms that only uses labeled data a large margin, consistently.} \label{tab:unlabeled}
\end{center}
\end{table}}

{\flushleft\textbf{The influence of semi parameter $\lambda_u$. }}$\lambda_u$ controls the supervised proportion of unlabeled data. We compare the counting accuracy for using different $\lambda_u$ during training. The result is shown in Table~\ref{tab:parameter}.

All experiments are conducted on ShanghaiTech A with a labeled ratio of $5\%$ and we choose previous state-of-the-art GP~\cite{sindagi2020learning} for comparison. 
In general, when tuning the unlabeled parameter, the final performance has a small fluctuation and is consistently better than previous method. 
In specific, when $\lambda_u=1$ and $N_u=19N_l$, due to the lack of explicit label supervision, the accuracy is not satisfactory. And when $\lambda_u \leq 0.01$, 
the unlabeled loss is too small to fully help the supervision, 
so the performance also drops, which is almost on par with only using labeled data. 
During other experiments, we will choose $\lambda_u=0.1$.

\renewcommand{\tabcolsep}{10 pt}{
\begin{table}[t]
\small
\begin{center}
\begin{tabular}{c|c|c|c|c|c}
  \toprule[1pt]
  $\lambda_u$ & 0 & 0.001 & 0.005 & 0.01 & 0.05 \\
  \hline
  MAE & 94.1 & 92.5 & 90.4 & 91.2 & 86.5 \\
  MSE & 142.9 & 140.6 & 141.3 & 142.0 & 139.3 \\
  \toprule[1pt]
   $\lambda_u$ & 0.1 & 0.5 & 1 & \multicolumn{2}{|c}{GP~\cite{sindagi2020learning}}\\
  \hline
  MAE & \textbf{85.4} & 87.9 & 94.2 & \multicolumn{2}{|c}{102.0}\\
  MSE & 134.5 & \textbf{134.3} & 145.8 & \multicolumn{2}{|c}{172.0}\\
  \toprule[1pt]
\end{tabular}
\caption{The influence of unlabeled parameter $\lambda_u$ on ShanghaiTech A (labeled ratio $5\%$).} 
\vspace{-8mm}
%Generally speaking, the proportion adjustment of unlabeled loss has little fluctuation on the final result. But it will affect the training effect when it is too small or too large.}
\label{tab:parameter}
\end{center}
\end{table}}

{\flushleft\textbf{The impact of distribution assumptions of matching probability}} is studied in Table~\ref{tab:distribution}. 
%We conduct experiments to study the impact of distribution assumptions in contrastive learning. 
In addition to the Laplace distribution, we also assume a Normal distribution where the matching probability subjects to $\hat{\omega_i} \sim \mathbf{Normal}\left (a_i, 0.5 \right ) = \frac{\sqrt{2}}{\sqrt{\pi}}\ e^{-2(d-a_i)^2}$. \XP{Different assumptions of matching probability have little effect on the final result and both can achieve good accuracy.}
%The comparison results on four datasets with a labeled ratio of $5\%$ are shown in Table~\ref{tab:distribution}.

As the results shown, adopting different distribution assumptions will have little effect on the final results. Both Laplace and Gaussian assumptions can achieve good accuracy and surpass previous state-of-the-art methods by a large margin on semi-supervised crowd counting. This \XP{suggests that the model is insensitive to the setting of} the contrastive weights $\omega$.
%will be robust to the contrastive weights $\omega$.
% This indicates that when there are large noises in dividing positive and negative pairs of contrastive learning, it will be helpful to appropriately reduce the weights of uncertain positive and negative examples, and the model will be robust to such weights.

% \def\arraystretch{1.1}
\renewcommand{\tabcolsep}{10 pt}{
\begin{table}[t]
\small
\begin{center}
\begin{tabular}{c|cc|cc}
  \toprule[1pt]
  \multicolumn{1}{c}{\multirow{2}*{Datasets}} & \multicolumn{2}{c}{Laplace} &  \multicolumn{2}{c}{Normal}\\
  \multicolumn{1}{c}{} &  \multicolumn{1}{c}{MAE} & \multicolumn{1}{c}{MSE} & \multicolumn{1}{c}{MAE} & \multicolumn{1}{c}{MSE} \\
  \midrule
  UCF-QNRF & 120.2 & 209.3 & 117.3  & 201.1 \\
  JHU++ & 82.2 & 294.9 & 83.2 & 303.7 \\
  ShanghaiTech A & 85.4 & 134.5 & 85.9 & 135.3 \\
  ShanghaiTech B & 12.6 & 22.8 & 12.8 & 23.3 \\
  \toprule[1pt]
\end{tabular}
\caption{The comparison between two distribution assumptions in contrastive learning (labeled ratio $5\%$).}
%Different assumptions of matching probability have little effect on the final result and both can achieve good accuracy.
\vspace{-9mm}
\label{tab:distribution}
\end{center}
\end{table}}

\section{Conclusion}
This paper proposes a novel approach for semi-supervised crowd counting and demonstrates the benefits of building regional supervision signals across images. It is achieved by a learnable density agency, which semantically connects the foreground features to the agents and rectifies the backbone feature extraction networks. A noise depression Bayesian loss is also introduced to mitigate the problem of annotation noises. The proposed agency mechanism provides rich supervised signals while the training pipeline \XP{with gradient-stop} ensures that only the most reliable supervision signals are propagated. 
% It does not need any external tasks or additional models such as segmentation.
\XP{DACount achieves excellent performance upon four challenging crowd counting datasets. Extensive experiments also demonstrate its} potential in reducing annotations efforts.  In future, we will apply the proposed agency based semi-supervised learning scheme to other semi-supervised learning tasks.

%This paper presents a novel approach for semi-supervised crowd counting. Our work shows the benefits of building supervision signals based on image regions. It is achieved by a learnable density agency, which targets on different supervisions for positive or negative instances under labeled or unlabeled data, achieving both sufficiency and reliability. We also improve the training pipeline by proposing a uncertainty-aware contrastive learning structure to further exploit density-wise information. And a noise depression Bayesian loss is introduced to mitigate the problem of annotation noises. Specifically, we do not need any external tasks or additional models such as segmentation. Extensive experiments have demonstrated the potential of DACount in reducing annotations efforts while achieving excellent performance upon four challenging crowd counting datasets.

\begin{acks}
    {This work is funded by the National Key Research and Development Project of China (2019YFB1312000), the National Natural Science Foundation of China (62076195 and U20B2052), the Fundamental Research Funds for the Central Universities (AUGA5710011522), and the GuangDong Basic and Applied Basic Research Foundation (2020B1515130004).}
\end{acks}

\bibliographystyle{ACM-Reference-Format}
\balance
\bibliography{acm}

\newpage
\appendix

\section{Ablation Study}
We hold more extensive experiments to study the influence of different parameters and settings. All experimental results are held in ShanghaiTech A with \emph{a labeled ratio of  $5\%$}. We choose previous state-of-the-art GP~\cite{sindagi2020learning} for comparison. The final performances of our model under different settings are consistently better than the previous method.

\subsection{\XP{The convergence speed}}

\begin{figure}[t]
\begin{center}
    \includegraphics[width=0.48\textwidth]{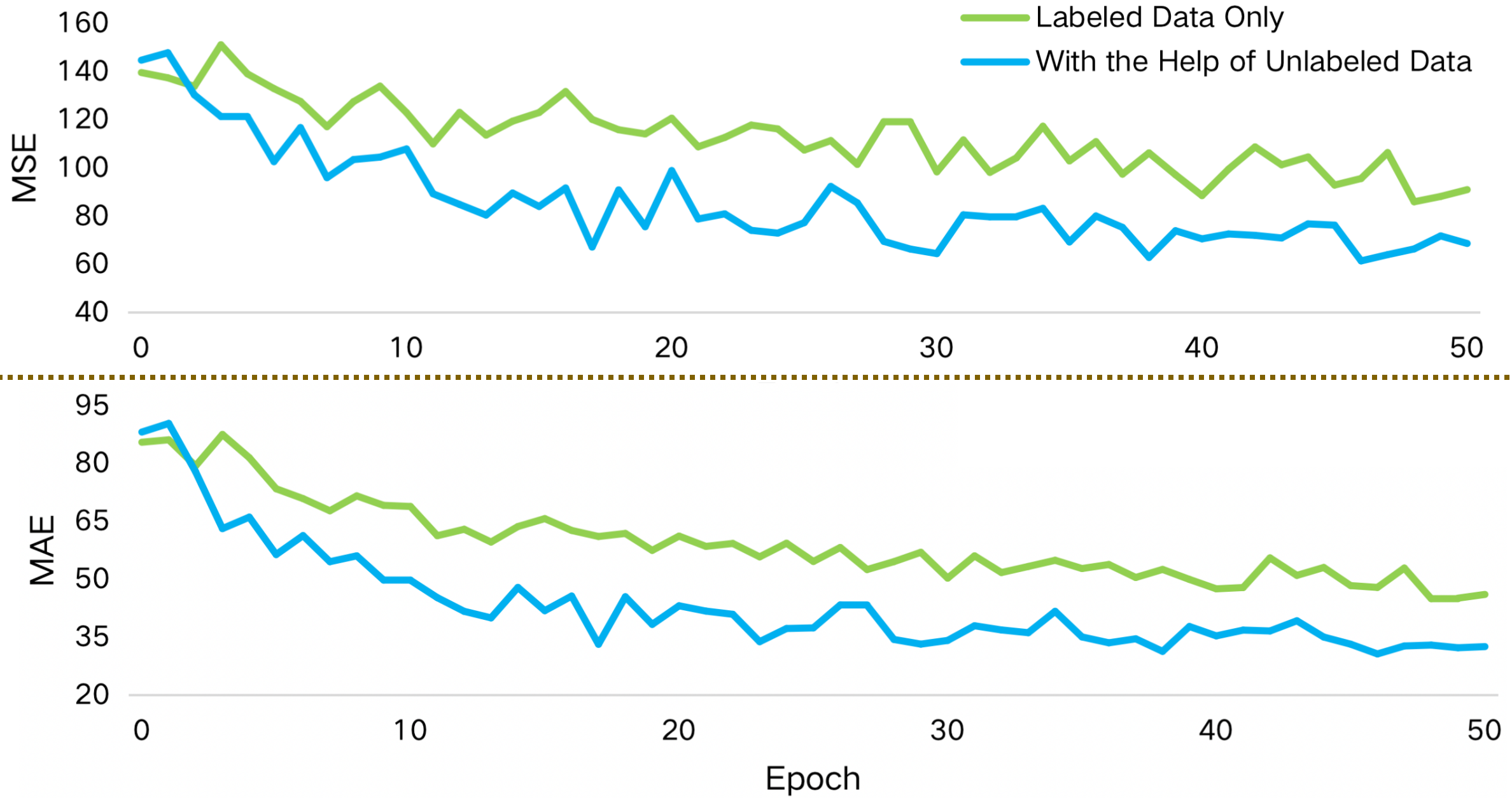}
\end{center}
\caption{The curves of training MAE and MSE under two different settings: training with only labeled data and training with both labeled and unlabeled data. With the help of unlabeled data, the accuracy of the model improves faster and performs consistently better.}
\label{fig:curve}
% \vspace{-4mm}
\end{figure}

To examine the impact of unlabeled data, we record the MAE and MSE of training epochs. The comparison result is shown in Figure~\ref{fig:curve}, where the green curve denotes the model trained only by labeled data and the blue one denotes the model trained by both labeled and unlabeled data.

For the first 20 epochs, we find that with the help of unlabeled data, the error in training samples drops faster than that of training with only labeled data. And after that, this low training error is maintained and consistently better. This quantitative experiment proves that our model can get great help from unlabeled data to improve the performance.

\subsection{The influence of noise depression}

We conduct experiments to study the influence of noise depression parameter $\beta$, which controls the strength of penalty. The comparison result is shown in Table~\ref{tab:beta}.

When $\beta = 0$, the loss degenerates
into normal Bayesian loss. And when we focus on $\beta > 0$, the results are constantly better than using all annotations indiscriminately. As $\beta$ is set larger, the depression of instances with large gaps will be more severe, but the final accuracy of the model becomes worse. It may suggest that when the depression strength is too strong, some appropriate supervision signals are also ignored, which leads to a deterioration of performance.

\def\arraystretch{1.1}
\renewcommand{\tabcolsep}{12 pt}{
\begin{table}[t]
\small
\begin{center}
\begin{tabular}{c|c|c|c|c}
  \toprule[1pt]
  
  $\beta$ & 0 & 0.1 & 0.5 & 1\\
  \hline
  MAE & 87.2 & 86.7 & 86.1 & \textbf{85.4} \\
  MSE & 140.8 & 136.7 & 137.6 & \textbf{134.5} \\
  \toprule[1pt]
   $\beta$ & 2 & 5 & 10 & GP~\cite{sindagi2020learning}\\
  \hline
  MAE & 87.1 & 87.2 & 87.0 & 102.0\\
  MSE & 134.8 & 136.2 & 138.9 & 172.0\\
  \toprule[1pt]
\end{tabular}
\caption{The influence of noise depression $\beta$. When $\beta$ is greater than $0$, the model accuracy can be improved. It indicates that reducing the weights of noisy supervisions is beneficial to the model.} \label{tab:beta}
\end{center}
\end{table}}

\subsection{The influence of density agency}
The parameter $\lambda_c$ determines how much that the density agency will supervise the model. Specifically, when $\lambda_c = 0$, the density agency will not work and the model is only supervised by ground-truth of labeled data. The experimental result is shown in Table~\ref{tab:lc}.

As it shown, without the help of density agency, the performance of the counting model drops a lot. And when $\lambda_c = 0.01$, the supervisions between density agency and the ground-truth achieves a good balance, where the model performs the highest accuracy. However, as $\lambda_c$ continues to increase, the proportion of this supervision becomes larger, gradually replacing the supervisions from ground-truth. The accuracy of the model gradually decreases, even approaching the performance of not using the agency.

\def\arraystretch{1.1}
\renewcommand{\tabcolsep}{12 pt}{
\begin{table}[t]
\small
\begin{center}
\begin{tabular}{c|c|c|c|c}
  \toprule[1pt]
  $\lambda_c$ & 0 & 0.001 & 0.01 & 0.05\\
  \hline
  MAE & 94.2 & 85.5 & \textbf{85.4} & 85.7 \\
  MSE & 148.9 & 135.1 & \textbf{134.5} & 134.6 \\
  \toprule[1pt]
   $\lambda_c$ & 0.1 & 0.5 & 1 & GP~\cite{sindagi2020learning}\\
  \hline
  MAE & 88.9 & 89.9 & 92.6 & 102.0\\
  MSE & 136.4 & 135.9 & 139.9 & 172.0\\
  \toprule[1pt]
\end{tabular}
\caption{The influence of agency loss parameter $\lambda_c$. With the moderate help of the density agency, the accuracy of the counting model has been significantly improved.} \label{tab:lc}
\end{center}
\end{table}}

\subsection{The influence of temperature parameter}

We conduct experiments of the temperature parameter $\tau$ in density agent guided contrastive learning. When $\tau$ is larger, the difference between different density features is smaller. The supervision of contrastive learning will be weakened. The result is shown in Table~\ref{tab:tau}.

In general, when tuning $\tau$, the final performance does not fluctuate much and is consistently quite better than the previous method. Particularly, when $\tau$ is too large or too small, the accuracy will still be affected to a certain extent. During other experiments, we keep $\tau=0.1$.

\def\arraystretch{1.1}
\renewcommand{\tabcolsep}{12 pt}{
\begin{table}[t]
\small
\begin{center}
\begin{tabular}{c|c|c|c|c}
  \toprule[1pt]
  $\tau$ & 0.01 & 0.05 & 0.07 & 0.1\\
  \hline
  MAE & 89.6 & 87.8 & 86.0 & \textbf{85.4} \\
  MSE & 138.4 & 136.3 & 135.8 & \textbf{134.5} \\
  \toprule[1pt]
   $\tau$ & 0.2 & 0.5 & 1 & GP~\cite{sindagi2020learning}\\
  \hline
  MAE & 85.4 & 86.7 & 88.9 & 102.0\\
  MSE & 134.7 & 135.6 & 139.2 & 172.0\\
  \toprule[1pt]
\end{tabular}
\caption{The influence of temperature parameter. In general, the adjustment of contrastive temperature has little fluctuation on the final result.} \label{tab:tau}
\end{center}
\end{table}}

\subsection{The influence of mask loss}
The mask loss parameter $\lambda_m$ controls the weight of the supervision targeted on foreground predictor. We conduct experiments to study the influence of mask loss, which result is shown in Table~\ref{tab:lm}.

Since dividing foreground and background is important for subsequent transformer and other structures, setting a $\lambda_m$ that is too small is inappropriate. As the experiment shows, when $\lambda_m = 0.01$, the performance of the counting model deteriorates. However, when the weight is too large, that is $\lambda_m = 1$, the accuracy also drops. It can be explained as the model pays too much attention to the division of the foreground and background, while ignoring the difference between different foreground densities.

\def\arraystretch{1.1}
\renewcommand{\tabcolsep}{12 pt}{
\begin{table}[t]
\small
\begin{center}
\begin{tabular}{c|c|c|c|c|c}
  \toprule[1pt]
  $\lambda_m$ & 0.01 & 0.05 & 0.01 & 0.5 & 1\\
  \hline
  MAE & 90.8 & 87.9 & \textbf{85.4} & 85.8 & 90.5 \\
  MSE & 142.0 & 137.7 & 134.5 & \textbf{134.4} & 146.7 \\
  \toprule[1pt]
\end{tabular}
\caption{The influence of mask loss. When the loss weight $\lambda_m$ is too small or too large, the performance of the counting model will deteriorate.} \label{tab:lm}
\end{center}
\end{table}}



\end{document}